\documentclass{egpubl}
\usepackage{eurovis2025}
 
\SpecialIssuePaper         

\CGFccby

\usepackage[T1]{fontenc}
\usepackage{dfadobe}  

\usepackage{cite}  
\BibtexOrBiblatex

\electronicVersion
\PrintedOrElectronic
\ifpdf \usepackage[pdftex]{graphicx} \pdfcompresslevel=9
\else \usepackage[dvips]{graphicx} \fi

\usepackage{egweblnk} 

\usepackage{adjustbox}
\usepackage[dvipsnames]{xcolor}

\usepackage[font=footnotesize]{subcaption}
\usepackage{float}

\usepackage{todonotes}
\usepackage{xparse}
\usepackage{fancyhdr}
\usepackage{xurl}
\usepackage{enumitem}
\usepackage{environ}
\usepackage{xspace}
\usepackage{scalefnt}
\usepackage{comment}
\usepackage{multirow}

\usepackage[]{caption} 
\fancypagestyle{plain}{
	\fancyhf{} 
	\fancyfoot[C]{\thepage} 
}
\ifdefined\dochanges
  \definecolor{changedcol}{RGB}{145, 42, 120}
  \definecolor{changedcolTVCG}{RGB}{145, 42, 120}
\else
  \definecolor{changedcol}{rgb}{0., 0., 0.}
  \definecolor{changedcolTVCG}{RGB}{0., 0., 0.}
\fi
\DeclareDocumentCommand{\changedFrom}{}{\color{changedcol}}
\DeclareDocumentCommand{\changedTo}{}{\color{black}}

\newenvironment{new}[1]{\color{changedcol}#1}{\color{black}}

\makeatletter
\newcommand{\cmmnt}[1]{\@bsphack\@esphack}
\makeatother
\environbodyname\envbody
\NewEnviron{old}{\cmmnt{\envbody}}

\newcommand{\Dhat}{{\hat D}}
\newcommand{\Fts}[1]{{\hat F^{#1}_{t\rightarrow s}}}
\newcommand{\Ftu}[1]{{\hat F^{#1}_{t\rightarrow u}}}

\newcommand{\Ft}[1]{{\hat F^{\,#1}_{t}}}
\newcommand{\Dts}[1]{{\hat D^{#1}_{t\leftarrow s}}}
\newcommand{\Dtu}[1]{{\hat D^{#1}_{t\leftarrow u}}}

\newcommand{\Fhat}{{\hat F}}

\newcommand{\highleftarrow}{\hbox{\raise.80ex\hbox{$\mathbin{^\leftarrow}$}}}
\newcommand{\W}{\hbox{$\mathbin{W\mskip-15.0mu\highleftarrow}$}}

\newcommand{\Loss}{\mathcal{L}}
\newcommand{\Lrec}{\mathcal{L}_{rec}}

\newcommand{\DGT}{D^{GT}_t}
\newcommand{\FGT}{F^{GT}_t}

\newcommand{\cb}[1]{\mbox{Conv Block$^{#1}$}}

\newcommand{\acronym}{HyperFLINT\xspace}

\usepackage{tikz}
\usetikzlibrary{shapes.geometric, arrows, calc}
\usetikzlibrary{positioning, shapes.multipart, arrows.meta}
\usetikzlibrary{arrows,shapes,automata,backgrounds,petri,positioning}
\usetikzlibrary{decorations.pathmorphing}
\usetikzlibrary{decorations.shapes}
\usetikzlibrary{decorations.text}
\usetikzlibrary{decorations.fractals}
\usetikzlibrary{decorations.footprints}
\usetikzlibrary{shadows}
\usetikzlibrary{calc}
\usetikzlibrary{spy}

\graphicspath{{figures/}{pictures/}{images/}{./}} 

\usepackage{microtype}                 
\PassOptionsToPackage{warn}{textcomp}  
\usepackage{textcomp}                  
\usepackage{mathptmx}                  
\usepackage{times}                     
\usepackage{cite}                      
\usepackage{tabu}                      
\usepackage{booktabs}                  
\usepackage{amsmath}
\usepackage{amssymb}

\newcommand{\orangecircle}{\tikz{\fill[orange] (0,0) circle (2pt);}}
\newcommand{\bluecircle}{\tikz{\fill[blue] (0,0) circle (2pt);}}
\newcommand{\greencircle}{\tikz{\fill[green] (0,0) circle (2pt);}}

\usepackage{fancyhdr}
\title{\acronym: Hypernetwork-based Flow Estimation and Temporal Interpolation for Scientific Ensemble Visualization}

\author[H. Gadirov, Q. Wu, D. Bauer, K.-L. Ma, J. B. T. M. Roerdink, \& S. Frey]
{\parbox{\textwidth}{\centering Hamid Gadirov$^{1}$\orcid{0000-0001-6578-4342}, 
Qi Wu$^{3}$\orcid{0000-0003-0342-9366}, 
David Bauer$^{2}$\orcid{0000-0002-1327-3054}, 
Kwan-Liu Ma$^{2}$\orcid{0000-0001-8086-0366}, 
Jos B.T.M. Roerdink$^{1}$\orcid{0000-0003-1092-9633}, 
and Steffen Frey$^{1}$\orcid{0000-0002-1872-6905}}
        \\
{\parbox{\textwidth}{\centering $^1$University of Groningen, Netherlands\\
         $^2$University of California, Davis, USA\\
         $^3$NVIDIA, USA}
}
}

\begin{document}


\maketitle

\begin{abstract}
  We present \acronym (Hypernetwork-based {FL}ow estimation and temporal {INT}erpolation), a novel deep learning-based approach for estimating flow fields, temporally interpolating scalar fields, and facilitating parameter space exploration in spatio-temporal scientific ensemble data.
  This work addresses the critical need to explicitly incorporate ensemble parameters into the learning process, as traditional methods often neglect these, limiting their ability to adapt to diverse simulation settings and provide meaningful insights into the data dynamics.
  \acronym introduces a hypernetwork to account for simulation parameters, enabling it to generate accurate interpolations and flow fields for each timestep by dynamically adapting to varying conditions, thereby outperforming existing parameter-agnostic approaches. The architecture features modular neural blocks with convolutional and deconvolutional layers, supported by a hypernetwork that generates weights for the main network, allowing the model to better capture intricate simulation dynamics. A series of experiments demonstrates \acronym's significantly improved performance in flow field estimation and temporal interpolation, as well as its potential in enabling parameter space exploration, offering valuable insights into complex scientific ensembles.


\end{abstract}  
\section{Introduction}
\vspace{-3pt}

Advancements in simulation and capture technologies now enable the observation of time-dependent processes at extremely high spatial and temporal resolutions. This generates vast spatio-temporal datasets, offering researchers opportunities to analyze parameter variations and stochastic effects. However, the sheer data volumes often exceed storage capacities, requiring selective preservation of timesteps or variables~\cite{childs2019situ}. Additionally, experimental data is constrained by specific modalities, limiting its comprehensiveness.
Addressing these challenges, several recent methods were proposed for reconstructing scalar fields~\cite{han2022coordnet, wu2023hyperinr, tang2024stsr, gadirov2024flint} and estimating flow fields~\cite{gadirov2024flint} in 2D and 3D scientific data.
FLINT's~\cite{gadirov2024flint} student-teacher architecture achieved state-of-the-art results in density interpolation and flow estimation, proving effective for reconstructing missing data. 
This is particularly valuable for applications like flow visualization~\cite{janicke_visual_2011}, optimal timestep selection~\cite{frey_flow-based_2017}, and ensemble member comparisons~\cite{tkachev_local_2021}.
However, FLINT's and the above-mentioned recent methods' limitations in generalizing across parameter space and neglecting ensemble parameters reduced their effectiveness in dynamic simulation scenarios.

\begin{figure}[t]
\centering
\begin{tikzpicture}[>=stealth,font=\sffamily]
    \tikzstyle{input} = [rectangle, draw, fill=blue!20,     text width=7.25em, text centered, minimum height=2.5em]
    \tikzstyle{nn} = [rectangle, draw, fill=orange!40,     text width=6em, text centered, rounded corners, minimum height=8em]
    \tikzstyle{hypernn} = [rectangle, draw, fill=red!40, text width=6em, text centered, rounded corners, minimum height=3em]
    \tikzstyle{output} = [rectangle, draw, fill=green!20,     text width=7.25em, text centered, minimum height=2.8em]
    
    \node [input] (input1) {Scalar field $D_s$ {\tiny (density, luminance)}};
    \node [input, below of=input1, yshift=-0.15cm] (input2) {Scalar field $D_u$ {\tiny (density, luminance)}};
    \node [input, above of=input1, yshift=0.75cm] (params) {Simulation Parameters};

    \node [hypernn, right of=params, xshift=2.0cm] (hypernet) {HyperNet};
    \node [nn, right of=input1, xshift=2.0cm, yshift=-0.5cm] (nn) {FLINT*\\ network};
    
    \node [output, right of=nn, xshift=2.0cm, yshift=0.5cm, align=center] (output1) {Flow field $\hat{F}_{t}$};
    \node [output, below of=output1, yshift=-0.15cm, align=center] (output2) {Scalar field $\hat{D}_{t}$ {\tiny (density, luminance)}};
    
    \draw [->, line width=1.5pt] (1.35, -0.6) -- (1.9, -0.6);
    \draw [->, line width=1.5pt] (params) -- (hypernet);
    \draw [->, line width=1.5pt] (hypernet) -- node[midway, right, font=\small] {$\theta$} (nn);
    \draw [->, line width=1.5pt] (4.15, -0.6) -- (4.7, -0.6);

\end{tikzpicture}
\vspace{-15pt}
\caption{Overview of \acronym pipeline during inference. The FLINT* deep neural network, whose weights are generated by the HyperNet, performs flow field estimation $\hat{F}_{t}$ and temporal (scalar) field interpolation $\hat{D}_{t}$, where $s < t < u$, by utilizing the available densities $D_s$ and $D_u$ from the previous and following timesteps, and their simulation parameters. 
}
\label{fig:nn_overview}
\vspace{-15pt}
\end{figure}
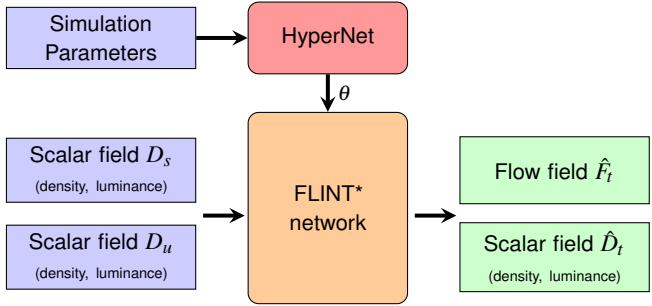

\vspace{-3pt}
In this paper, we introduce \acronym (\underline{Hyper}network-based \underline{FL}ow estimation and temporal \underline{INT}erpolation), a deep learning approach that uses hypernetworks to estimate missing flow fields in scientific ensembles. \acronym leverages the simulation parameters via hypernetworks to generate accurate flow fields for each timestep, even in scenarios where the flow could not be captured or was omitted \changedFrom{}(see overview in~\autoref{fig:nn_overview})\changedTo{}.
Additionally, \acronym is capable of producing high-quality temporal interpolants between scalar fields, providing a comprehensive solution for reconstructing both flow and scalar data.
\acronym improves upon existing state-of-the-art methods, such as FLINT~\cite{gadirov2024flint}, by focusing on scientific ensembles and incorporating hypernetworks to account for simulation parameters, resulting in better temporal interpolation and flow estimation. Unlike previous approaches, \acronym handles complex spatio-temporal datasets without requiring domain-specific assumptions, pre-training, or fine-tuning on simplified datasets.

The ability of \acronym to incorporate parameter-driven transformations into its model architecture opens up new possibilities for parameter space exploration. By learning the relationships between simulation parameters and output data,
\acronym can generate approximations of data for configurations that were not explicitly simulated, allowing for a broader investigation of possible simulation outcomes without rerunning the full simulation for each parameter set. This capacity is particularly valuable for understanding complex dependencies between parameters in ensemble simulations, as it enables researchers to interpolate or extrapolate within the parameter space. With \acronym, users can systematically explore how changes in parameters impact the generated fields, offering insights into the underlying phenomena that would otherwise require extensive computational resources to simulate directly.
Moreover, \acronym's parameter-driven capabilities allow for the reconstruction of missing data, including variables and timesteps, which is especially useful in large-scale simulations where only a subset of the data can be saved~\cite{childs2020terminology}. 
The \acronym code will be made available publicly.

Our key contributions can be summarized as follows:

\begin{enumerate}[leftmargin=*, label=\textbullet, noitemsep] 
\item To the best of our knowledge, \acronym is the first approach that employs a hypernetwork to adaptively estimate flow fields and produce state-of-the-art temporal interpolations of density fields for scientific ensembles by dynamically conditioning on simulation parameters.


\item \acronym effectively handles spatio-temporal ensembles utilizing a hypernetwork, requiring no specific assumptions, 
making it versatile for diverse scientific applications. 
\item With the introduction of the hypernetwork, \acronym achieves several key advancements: it dynamically generates simulation parameter-aware weights facilitating the understanding of spatio-temporal scientific ensembles, enhancing model quality and performance by adapting to varying simulation parameters, enabling significantly improved flow estimation and scalar field interpolation, even in scenarios with sparse or incomplete data. 


\end{enumerate}


\vspace{-6pt}
\section{Related Work}
\label{sec:related_work}

\textbf{Hypernetworks in deep learning.} 
Hypernetworks~\cite{ha2016hypernetworks} have gained traction in the deep learning community as an effective mechanism for generating the weights of another neural network, thereby offering greater flexibility in capturing complex relationships between data and model parameters. 
These networks excel in tasks where parameter space exploration is crucial, as they can dynamically adjust the parameters of a target network based on external inputs, such as simulation parameters. 
Hypernetworks have been successfully applied in various fields, including neural architecture search~\cite{brock2017smash}, generative modeling~\cite{skorokhodov2021adversarial}, and meta-learning~\cite{przewiezlikowski2022hypermaml}. 

\vspace{-3pt}
\textbf{Flow field estimation in scientific visualization.}
Several methods have been developed for flow field estimation and visualization in scientific datasets. Kappe \emph{et al.}\cite{kappe2015reconstruction} focused on estimating 3D local flow in microscopy data using a combination of image processing and optical flow techniques, while Kumpf \emph{et al.}\cite{kumpf2018visual} applied ensemble sensitivity analysis with optical flow-based feature tracking to study changes in geo-spatial data. Manandhar \emph{et al.}~\cite{manandhar2018sparse} proposed dense 3D optical flow estimation for microscopy image volumes using displacement vectors. However, these approaches often rely on specific assumptions, limiting their applicability to real-world scientific ensembles.
In contrast, \acronym operates without dataset-specific assumptions, making it versatile for diverse scientific visualization tasks. 
Sahoo \emph{et al.}\cite{sahoo2022integration} introduced Integration-Free Learning of Flow Maps, which estimates flow directly from state observations, enhancing efficiency for large-scale datasets where ground-truth (GT) flow may not be available.
FLINT~\cite{gadirov2024flint} emerged as a state-of-the-art approach for reconstructing scalar fields and simultaneously estimating flow fields in spatio-temporal scientific datasets. 
FLINT's student-teacher architecture and flexible loss function helped achieve high accuracy in density interpolation and flow estimation tasks, particularly in 2D+time and 3D+time datasets.
FLINT’s modular design features a series of convolutional and deconvolutional layers grouped into neural blocks, which iteratively refine scalar field and flow field outputs. Temporal consistency is enforced through specific loss components, ensuring alignment with GT data and improving interpolation accuracy. These capabilities made FLINT particularly effective for addressing missing data in large-scale simulations.
However, FLINT exhibits limitations in dynamic simulation scenarios where varying ensemble parameters affect data behavior. Without mechanisms to explicitly incorporate these parameter variations, FLINT’s adaptability and generalization remain constrained.

\vspace{-3pt}
\textbf{Machine learning-based upscaling \& super-resolution.} 
Across multiple fields, including image processing, computer vision, and scientific visualization~\cite{ledig2017photo, shi2016real}, ML-based upscaling and super-resolution approaches have gained considerable attention.
These techniques generally target either spatial, temporal, or combined spatio-temporal enhancements, forming three main categories: spatial super-resolution (SSR), temporal super-resolution (TSR), and spatio-temporal super-resolution (STSR).
SSR approaches, exemplified by models such as SRCNN~\cite{dong2015image}, SRFBN~\cite{li2019feedback}, and SwinIR~\cite{liang2021swinir}, focus on increasing spatial detail by generating realistic textures and enhancing fine structures. TSR methods, on the other hand, aim to fill in intermediate frames in time-subsampled sequences without degrading spatial quality. Methods like phase-based interpolation~\cite{meyer2015phase}, SepConv~\cite{niklaus2017video}, and SloMo~\cite{jiang2018super} are representative TSR techniques that focus on temporal resolution improvements in videos.
While some prior approaches, such as STNet~\cite{han2021stnet}, address either spatial or temporal resolution, they typically do not tackle both simultaneously. For example, TSR-TVD~\cite{han2019tsr}, a recurrent generative network proposed by Han \emph{et al.}, was designed to temporally upscale a variable different from the given one,
without explicitly addressing both spatial and temporal dimensions. 
\begin{new}
Volume scene representation networks (V-SRN)~\cite{lu2021compressive, park2019deepsdf, sitzmann2019scene, mildenhall2021nerf} have significantly advanced neural representations for volumetric data, enabling high-quality rendering and reconstruction through implicit neural representations. 
Building on these advancements, fV-SRN~\cite{weiss2022fast} leverages GPU tensor cores to integrate neural reconstruction into on-chip ray tracing kernels, reducing computational complexity and accelerating training and inference for real-time volume rendering.\end{new}
Recent advancements, such as Filling the Void~\cite{mishra2022filling}, SSR-TVD~\cite{han2020ssr}, FFEINR~\cite{jiao2023ffeinr}, HyperINR~\cite{wu2023hyperinr}, CoordNet~\cite{han2022coordnet}, and STSR-INR~\cite{tang2024stsr}, have shown improvements in handling either TSR or SSR of data fields at arbitrary resolutions.
FLINT~\cite{gadirov2024flint} can perform temporal interpolation for both 2D+time and 3D+time datasets but does not consider the parameter space of scientific ensembles, which limits its adaptability to different scenarios. 
\begin{new}
In contrast, \acronym introduces hypernetworks to incorporate ensemble parameters directly into the interpolation process, enabling a more adaptive and data-driven approach. 
\end{new}


\begin{figure*}[t]

\begin{tikzpicture}[>=stealth]

\node[rectangle, draw, fill=blue!30, text width=5em, text centered, minimum height=2.5em] (params) {{\textit{Parameters \tiny (simulation)}}};

\node[draw=red!40, minimum width=12.9cm, minimum height=3cm, right=1.0cm of params, line width=3pt, rounded corners] (hypernet) {};

\node[right=0.15cm of hypernet, yshift=0.9cm] {\textbf{\normalsize{HyperNet}}};

\node[draw, fill=orange!50, rounded corners, minimum width=0.5cm, minimum height=1.2cm, right=1.5cm of params] (mlp1) {\rotatebox{90}{\scriptsize Linear\\(\#params, 16)}};
\node[draw, fill=green!50, rounded corners, minimum width=0.5cm, minimum height=1.2cm, right=0.3cm of mlp1] (prelu1) {\rotatebox{90}{\scriptsize PReLU}};
\node[draw, fill=cyan!30, rounded corners, minimum width=0.5cm, minimum height=1.2cm, right=0.3cm of prelu1] (dropout1) {\rotatebox{90}{\scriptsize Dropout}};
\node[draw, fill=orange!50, rounded corners, minimum width=0.5cm, minimum height=2.1cm, right=0.3cm of dropout1] (mlp2) {\rotatebox{90}{\scriptsize Linear\\(16, 32)}};
\node[draw, fill=green!50, rounded corners, minimum width=0.5cm, minimum height=1.2cm, right=0.3cm of mlp2] (prelu2) {\rotatebox{90}{\scriptsize PReLU}};
\node[draw, fill=cyan!30, rounded corners, minimum width=0.5cm, minimum height=1.2cm, right=0.3cm of prelu2] (dropout2) {\rotatebox{90}{\scriptsize Dropout}};
\node[draw, fill=orange!50, rounded corners, minimum width=0.5cm, minimum height=1.2cm, right=0.3cm of dropout2] (mlp3) {\rotatebox{90}{\scriptsize Linear\\(32, 3)}};
\node[draw, fill=green!50, rounded corners, minimum width=0.5cm, minimum height=1.2cm, right=0.3cm of mlp3] (prelu3) {\rotatebox{90}{\scriptsize PReLU}};

\draw[dashed] ($(mlp1.north west)+(-0.25,0.23)$) rectangle ($(prelu3.south east)+(0.3,-0.57)$);
\node[below] at ($(mlp1.south)!0.5!(prelu3.south) + (0,-0.34)$) {\scriptsize MLP};

\node[draw, fill=red!50, rounded corners, minimum width=0.5cm, minimum height=1.2cm, right=0.8cm of prelu3] (conv1) {\rotatebox{90}{\scriptsize Conv1D\\(1, 16)}};
\node[draw, fill=green!50, rounded corners, minimum width=0.5cm, minimum height=1.2cm, right=0.4cm of conv1] (prelu4) {\rotatebox{90}{\scriptsize PReLU}};
\node[draw, fill=red!50, rounded corners, minimum width=0.4cm, minimum height=1.2cm, right=0.4cm of prelu4] (conv2) {\rotatebox{90}{\scriptsize Conv1D\\(16, 32)}};
\node[draw, fill=green!50, rounded corners, minimum width=0.5cm, minimum height=1.2cm, right=0.4cm of conv2] (prelu5) {\rotatebox{90}{\scriptsize PReLU}};
\node[draw, fill=gray!30, rounded corners, minimum width=0.5cm, minimum height=1.2cm, right=0.4cm of prelu5] (flatten) {\rotatebox{90}{\scriptsize Flatten}};
\node[draw, fill=blue!20, rounded corners, minimum width=0.5cm, minimum height=1.2cm, right=0.4cm of flatten] (linear2) {\rotatebox{90}{\scriptsize Linear\\(32$\times$3, \\kernels)}};

\draw[dashed] ($(conv1.north west)+(-0.3,0.45)$) rectangle ($(linear2.south east)+(0.2,-0.15)$);
\node[below] at ($(conv1.south)!0.5!(flatten.south) + (0.4,-0.44)$) {\scriptsize CNN};

\draw [->, line width=1.2pt] (1.1, -0.0) -- (1.8, -0.0);

\draw[->, thick] (hypernet.south) -- ++(0,-0.4) node[right] {$\theta$};

\draw[->, thick] (mlp1.east) -- (prelu1.west);
\draw[->, thick] (prelu1.east) -- (dropout1.west);
\draw[->, thick] (dropout1.east) -- (mlp2.west);
\draw[->, thick] (mlp2.east) -- (prelu2.west);
\draw[->, thick] (prelu2.east) -- (dropout2.west);
\draw[->, thick] (dropout2.east) -- (mlp3.west);
\draw[->, thick] (mlp3.east) -- (prelu3.west);

\draw[->, thick] (prelu3.east) -- (conv1.west);

\draw[->, thick] (conv1.east) -- (prelu4.west);
\draw[->, thick] (prelu4.east) -- (conv2.west);
\draw[->, thick] (conv2.east) -- (prelu5.west);
\draw[->, thick] (prelu5.east) -- (flatten.west);
\draw[->, thick] (flatten.east) -- (linear2.west);


\end{tikzpicture}
\label{fig:hypernet}
\vspace{-10pt}
\end{figure*}

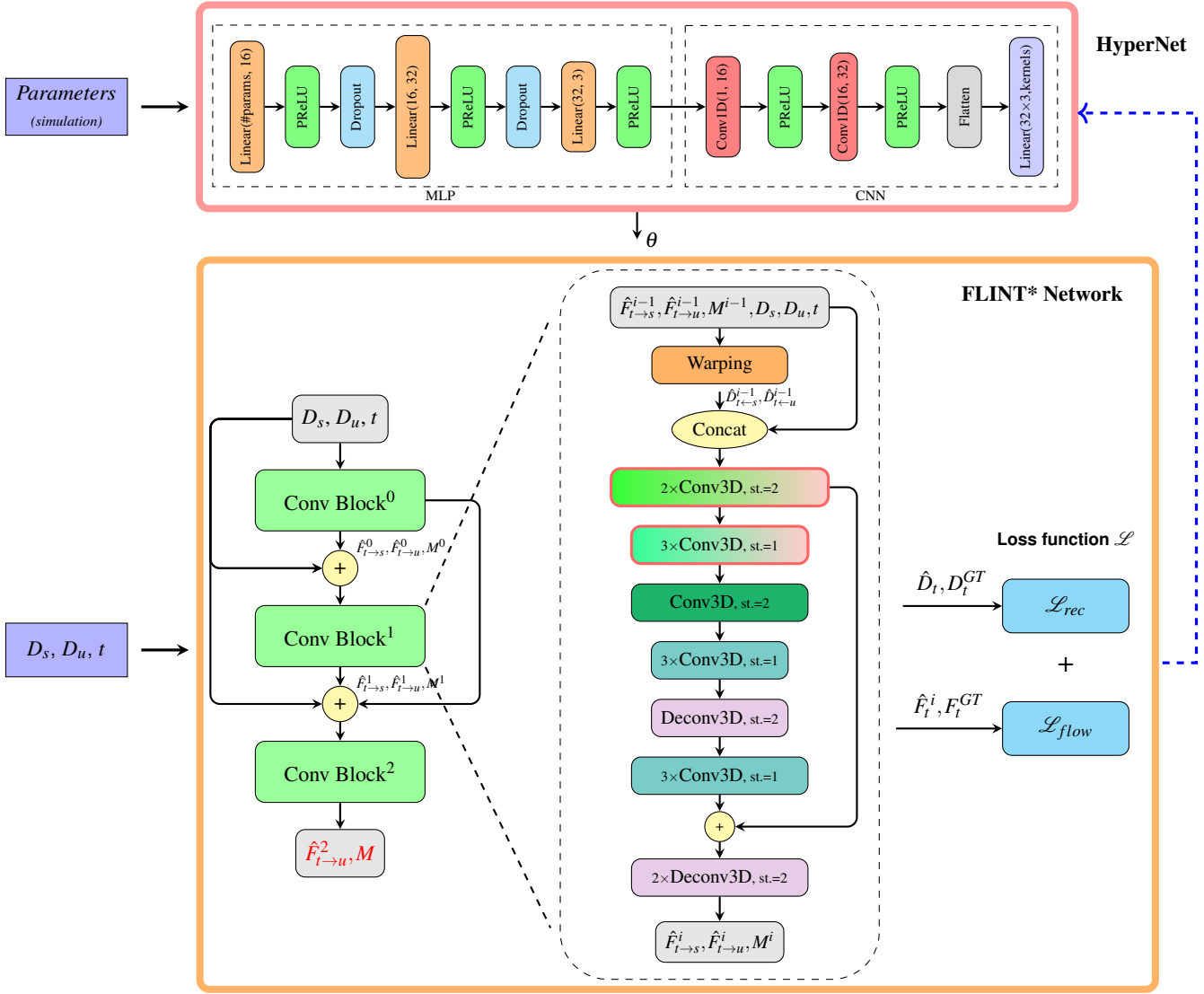
\begin{figure*}[t]
  \begin{adjustbox}{raise=4.65cm}
    \begin{minipage}[b]{0.15\textwidth}
      \centering
      \begin{tikzpicture}[>=stealth,font=\sffamily]
        \tikzstyle{input} = [rectangle, draw, fill=blue!30, text
        width=5em, text centered, minimum height=2.5em]
					
        \node (input1) [input] {$D_s$, $D_{u}$, $t$};
					
        \draw [->, line width=1.2pt] (1.1, -0.0) -- (1.8, -0.0);
					
      \end{tikzpicture}
    \end{minipage}
  \end{adjustbox}
	%
  \setlength{\fboxrule}{4pt}
  \begin{tikzpicture}
    \node[draw, orange!60, fill=white, rounded corners, inner sep=4pt,
    line width=3pt, text=black] at (0,0) {
      \begin{adjustbox}{raise=0.75cm}
      
        \begin{minipage}[b]{0.28\textwidth} %
          \tikzstyle{convblock} = [rectangle, rounded corners, minimum
          width=2.5cm, minimum height=0.9cm,text centered, draw=black,
          fill=green!40] \tikzstyle{arrow} = [thick,->,>=stealth]
          \begin{tikzpicture}[node distance=0.5cm]
	\node (input) [draw, fill=gray!20, minimum width=1.0cm, minimum height=0.7cm] {$D_s$, $D_{u}$, $t$};
	\node (conv0) [convblock, below of=input, yshift=-0.7cm] {Conv Block$^0$};
        \node (sum0) [draw, circle, below of=conv0, yshift=-0.5cm, fill=yellow!40, inner sep=0.1cm] {{+}};
        \node [above right of=sum0, font=\scriptsize, xshift=0.55cm, yshift=-0.05cm] {$\Fts{0}, \Ftu{0}, M^{0}$};
	\node (conv1) [convblock, below of=sum0, yshift=-0.5cm] {Conv Block$^1$};
	\node (sum1) [draw, circle, below of=conv1, yshift=-0.5cm, fill=yellow!40, inner sep=0.1cm] {{+}};
         \node [above right of=sum1, font=\scriptsize, xshift=0.55cm, yshift=-0.05cm] {$\Fts{1}, \Ftu{1}, M^{1}$};
	\node (conv2) [convblock, below of=sum1, yshift=-0.5cm] {Conv Block$^{2}$};
	\node (output) [draw, below of=conv2, fill=gray!20, minimum width=1cm, minimum height=0.7cm, yshift=-0.7cm] {\color{red}$\Ftu{2}, M$};
						
	\draw [arrow] (input) -- (conv0);
	\draw [arrow] (conv0) -- (sum0);
	\draw [arrow] (sum0) -- (conv1);
	\draw [arrow] (conv1) -- (sum1);
        \draw [arrow] (sum1) -- (conv2);

        \draw [arrow] (conv2) -- (output);
	\draw [arrow] (input.west) -- ++(-1.2,0) |- (sum0.west);
	\draw [arrow] (input.west) -- ++(-1.2,0) |- (sum1.west);
	\draw [arrow] (input.west) -- ++(-1.2,0) |- (sum0.west);
	\draw [arrow] (conv0.east) -- ++(0.8,0) |- (sum1.east);

	\draw [thick, dashed] (conv1.north east) -- (3.1, 1.5);
	\draw [thick, dashed] (conv1.south east) -- (3.1, -7.5);
      \end{tikzpicture}
    \end{minipage}
  \end{adjustbox}
  
  \begin{minipage}[b]{0.28\textwidth} 
    \color{black} \centering
    \tikzstyle{convlayer} = [rectangle, rounded corners, minimum width=3.5cm, minimum height=0.55cm,text centered, draw=black, fill=Emerald!50]
    \tikzstyle{warp} = [rectangle, rounded corners, minimum width=2.0cm, minimum height=0.55cm,text centered, draw=black]
    \tikzstyle{concat} = [ellipse, minimum width=0.5cm, minimum height=0.55cm,text centered, draw=black, fill=green!20]
    \tikzstyle{arrow} = [thick,->,>=stealth]
    \begin{tikzpicture}[node distance=0.45cm]
      \node (input) [draw, fill=gray!20, minimum width=1cm, minimum height=0.5cm, yshift=0.5cm, font=\footnotesize] {$\Fts{i-1}, \Ftu{i-1}, M^{i-1}, D_s, D_{u}, t$};
      \node (warp) [warp, below of=input, yshift=-0.4cm, fill=orange!60, font=\footnotesize] {Warping};
        \node [below right of=warp, font=\scriptsize, xshift=0.3cm, yshift=-0.15cm] {$\Dts{i-1}, \Dtu{i-1}$};
      \node (sum0) [concat, below of=warp, yshift=-0.5cm, fill=yellow!40, inner sep=0.1cm, font=\footnotesize] {Concat};
      \draw [arrow] (input.east) -- ++(0.4,0) |- (sum0.east);
\node (conv1) [convlayer, below of=sum0, yshift=-0.4cm, draw=red!60, very thick, minimum width=3.2cm, font=\scriptsize,
    shading=axis, shading angle=90, left color=Green!80, right color=red!20] {2$\times$\footnotesize{Conv3D}\scriptsize, st.=2};
    \node (conv2) [convlayer, below of=conv1, yshift=-0.4cm, draw=red!60, very thick, minimum width=2.6cm, font=\scriptsize, shading=axis, shading angle=90, left color=Emerald!80, right color=red!20] {3$\times$\footnotesize{Conv3D}\scriptsize, st.=1};
       \node (conv3) [convlayer, below of=conv2, yshift=-0.4cm, minimum width=2.6cm, fill=Green!80, font=\scriptsize] {\footnotesize{Conv3D}\scriptsize, st.=2}; 
       \node (conv4) [convlayer, below of=conv3, yshift=-0.4cm, minimum width=2.0cm, font=\scriptsize] {3$\times$\footnotesize{Conv3D}\scriptsize, st.=1};
        \node (conv5) [convlayer, below of=conv4, yshift=-0.4cm, minimum width=2.0cm, fill=violet!20, font=\scriptsize] {\footnotesize{Deconv3D}\scriptsize, st.=2}; 
        \node (conv6) [convlayer, below of=conv5, yshift=-0.4cm, minimum width=2.6cm, font=\scriptsize] {3$\times$\footnotesize{Conv3D}\scriptsize, st.=1};
	\node (sum2) [draw, circle, below of=conv6, yshift=-0.3cm, fill=yellow!40, inner sep=0.1cm, font=\scriptsize] {+};
	\node (conv7) [convlayer, below of=sum2, yshift=-0.3cm, minimum width=2.6cm, fill=violet!20, font=\scriptsize] {2$\times$\footnotesize{Deconv3D}\scriptsize, st.=2};
	\node (output) [draw, below of=conv7, fill=gray!20, minimum width=1cm, minimum height=0.5cm, yshift=-0.5cm, font=\footnotesize] {$\Fts{i}, \Ftu{i}, M^{i}$};
							
							\draw [arrow] (input) -- (warp);
							\draw [arrow] (conv0) -- (sum0);
							\draw [arrow] (sum0) -- (conv1);
							\draw [arrow] (conv1) -- (conv2);
                                \draw [arrow] (conv2) -- (conv3);
                                \draw [arrow] (conv3) -- (conv4);
                                \draw [arrow] (conv4) -- (conv5);
                                \draw [arrow] (conv5) -- (conv6);
							\draw [arrow] (conv6) -- (sum2);
							\draw [arrow] (sum2) -- (conv7);
							\draw [arrow] (conv7) -- (output);
							\draw [arrow] (conv1.east) -- ++(0.4,0) |- (sum2.east);

							\draw[dashed, rounded corners=20pt] (-2.35,-9.4) rectangle (2.35,1.05);
						\end{tikzpicture}%
				\end{minipage}
				\begin{adjustbox}{raise=3.3cm}
					\begin{minipage}[b]{0.21\textwidth} 
                    \color{black}
						\centering
						\begin{tikzpicture}[>=stealth,font=\sffamily]
							\tikzstyle{loss1} = [rectangle, draw, fill=cyan!40, text width=5em, text centered, minimum height=2.5em]
                                \tikzstyle{arrow} = [thick,->,>=stealth]
							\node [text=black, above] at (0,0.75) {\small\textbf{Loss function $\Loss$}};
							\node [loss1] (loss1) {$\mathcal{L}_{rec}$};
							\node [below=0.2cm of loss1, font=\Large] {+};
							
							\node [loss1, below=1cm of loss1] (loss2) {$\mathcal{L}_{flow}$};

                                \draw [arrow] (loss1.west -| {-2.4cm,0}) -- node[midway, above] {$\Dhat_{t}, {\DGT}$}  (loss1.west);
                                 
                               \draw [arrow] (loss2.west -| {-2.5cm,0}) -- node[midway, above] {$\Ft{i}, {\FGT}$}  (loss2.west);
							
						\end{tikzpicture}
					\end{minipage}
				\end{adjustbox}
       };

\end{tikzpicture}
		\begin{picture}(0,0)
			\put(-85, 290){\textbf{\normalsize{FLINT* Network}}}
		\end{picture}

\begin{tikzpicture}[remember picture, overlay]
    \draw[dashed, ->, line width=1.2pt, color=blue] 
        (17.05, 5.25) -- (17.55, 5.25) -- (17.55, 13.35) -- (15.8, 13.35); 
\end{tikzpicture}
\vspace{-18pt}
		\caption{ \acronym network architecture
                  and pipeline during training:
                  Given the
                  input fields $D_s$ and $D_{u}$, and their simulation parameters,
                  \acronym predicts the $\hat{D}_{t}$ scalar field
                  and $\hat{F}^{i}_{t}$ flow fields used in the loss
                    function for optimizing network parameters.  
                    The \acronym model consists of two key components: the HyperNet and the main network, FLINT*. 
                    The HyperNet, depicted within the red box, generates weights for the convolutional layers of FLINT* (with red outlines in the middle column of the orange box).
                    The FLINT* model architecture and loss function are shown in the orange box.  
                    The model consists of several stacked blocks of the
                  convolutional network, which takes $D_s$, $D_{u}$,
                  and $t$ as input and in the $i^{th}$ $\cb{}$ computes
                    estimated flows $\Fts{i}, \Ftu{i}$, and fusion
                    mask $M^i$ used for interpolation.  
                  The zoomed-in view on the right highlights the structure of a generic Conv Block.
                  The GT density $\DGT$ and flow $\FGT$ is only used in the loss function $\Loss$.
                  The blue dashed arrow in the right of the figure represents the gradient propagation during training, from the output of FLINT* back to the HyperNet. 
                    }
		\label{fig:model}
        \vspace{-12pt}
\end{figure*}




\vspace{-3pt}
\section{Method}
\vspace{-2pt}
\label{sec:method}
To handle diverse parameter settings in spatio-temporal scientific ensembles, we introduce \acronym, a method that integrates a hypernetwork to dynamically adapt the main neural network, FLINT*. The hypernetwork generates weights based on simulation parameters.
\begin{new}The pipeline consists of three main components: (1) the hypernetwork (HyperNet), which takes simulation parameters as input and produces weights for FLINT*, which is a simplified and streamlined version of the original FLINT network; (2) the FLINT* network, which estimates flow fields and interpolates scalar data across time steps; and (3) a training framework that optimizes both flow estimation and interpolation through a combination of loss functions. Unlike traditional methods, \acronym does not require pre-training or fine-tuning on simplified datasets, enabling efficient adaptation to new scenarios.\end{new}
The following sections elaborate on the neural network architecture of \acronym (\autoref{subsec:mehtod_net}), describe the temporal interpolation and flow estimation pipeline used in both training and inference (\autoref{sec:method:interp}), and detail our proposed loss function (\autoref{subsec:loss}).
\begin{new}
A detailed comparison between \acronym and FLINT can be found in \autoref{subsec:hyper_vs_flint}.
\end{new}

\begin{old}
To handle diverse parameter settings in spatio-temporal scientific ensembles, we introduce \acronym, a method that employs a hypernetwork to dynamically adapt the main neural network FLINT*. The hypernetwork generates weights conditioned on simulation parameters, enabling \acronym to flexibly interpolate temporal data and estimate flow fields accurately under varying conditions. This adaptability addresses challenges in scenarios where parameter variations impact data behavior, supporting improved modeling and analysis. To the best of our knowledge, \acronym is the first method to leverage hypernetworks for directly estimating flow fields from scalar data within scientific ensembles while achieving temporal super-resolution.
The hypernetwork design in \acronym allows for comprehensive exploration of parameter space, offering deeper insights into complex system dynamics. Unlike traditional methods that rely heavily on ground-truth flow during training, \acronym's hypernetwork-based approach can interpolate between scalar fields across time steps, making it especially suitable for scientific datasets where labeled flow data is often limited. Additionally, \acronym requires no pre-training or fine-tuning on simplified datasets, achieving rapid convergence on target datasets—reaching 92\% accuracy within the first 30\% of training time. This efficiency contrasts with computer vision approaches that often depend on extensive training on surrogate datasets~\cite{teed2020raft, dosovitskiy2015flownet, luo2021upflow}.
The following sections elaborate on the neural network architecture of \acronym (\autoref{subsec:mehtod_net}), detail the temporal interpolation and flow estimation pipeline used in both training and inference (\autoref{sec:method:interp}), and discuss our proposed loss function (\autoref{subsec:loss}).
\end{old}



\vspace{-3pt}
\subsection{\acronym Network Architecture}
\vspace{-2pt}
\label{subsec:mehtod_net}

The architecture of \acronym integrates two neural networks: HyperNet and FLINT*.
HyperNet (\autoref{fig:model}, top), described in \autoref{subsec:hypernet}, dynamically generates weights for the FLINT* network (\autoref{fig:model}, bottom), which is inspired by the FLINT method~\cite{gadirov2024flint}.
The architecture of the FLINT* network comprises $N = 3$ stacked convolutional blocks (\textit{Conv Block}), each incorporating convolutional (\textit{Conv}) and deconvolutional (\textit{Deconv}) layers.
The expanded view of a single convolutional block is presented in the middle column of the orange box in \autoref{fig:model}. 
\begin{new}
The first five \textit{Conv} layers (red outlines, middle column) dynamically adjust their weights based on the output from HyperNet.
\end{new}
The FLINT* network starts with 128 feature channels in the first block, reduces to 96 channels in the second block, and finishes with 64 channels in the final block.
PReLU activation~\cite{he2015delving} is applied to all layers except the last one, ensuring efficient learning and non-linear transformations.
The training process is driven by loss components illustrated on the right side of \autoref{fig:model}, guiding both flow estimation and scalar field interpolation. 
FLINT* and HyperNet are trained jointly, ensuring that the hypernetwork optimizes the weights of the main network dynamically throughout training.
\begin{old}
This integrated optimization helps \acronym achieve better flow estimation and temporal interpolation of scalar fields.
\end{old}

\vspace{-3pt}
\subsection{HyperNet}
\vspace{-2pt}
\label{subsec:hypernet}
The HyperNet architecture, as depicted in \autoref{fig:model} top part, is an integral part of \acronym. 
It takes numerical input parameters that characterize the simulation (such as physical quantities or configuration settings) and processes them through two main components: a Multi-Layer Perceptron (MLP) and a Convolutional Neural Network (CNN).
First, the MLP, consisting of three linear layers with PReLU activation functions and two dropout layers, transforms the input parameters into a higher-dimensional representation. This embedding allows the network to interpret the simulation parameters in a way that is beneficial for downstream tasks. 
Once processed by the MLP, the intermediate representation is passed into the CNN, where additional transformations are applied, refining the information derived from the parameters.
Finally, the output of the CNN is reshaped into a one-dimensional vector and passed through a final linear layer to generate the weights (\(\theta\)) for the convolutional layers of the FLINT* network.
The choice of one-dimensional convolutions (Conv1D) within the CNN component is motivated by the fact that the input simulation parameters are represented as one-dimensional sequences. Conv1D layers provide the most efficient way to capture local relationships between the parameters, preserving their spatial or sequential structure. Compared to fully connected layers, Conv1D can model these relationships in a way that is more computationally efficient, with fewer parameters, and can capture local patterns in parameters that are missed by dense layers. This structured approach helps the HyperNet leverage the inherent relationships between parameters to refine the generated weights.
The proposed architecture enables parameter-aware learning, where the simulation's governing parameters directly influence the model’s internal weights. This not only improves the model’s quantitative and qualitative performance but also facilitates parameter space exploration (\autoref{sec:param_space_exp}).
\begin{old}
By incorporating simulation parameters into the FLINT* network, HyperNet enables the model to adjust to different data generation conditions, leading to improved predictions across various scenarios.
\end{old}

\vspace{-3pt}
\subsection{Flow Estimation and Scalar Field Interpolation}
\vspace{-2pt}
\label{sec:method:interp}

\begin{new}
\acronym takes as input the simulation parameters (for HypetNet), two scalar fields $D_s$ and $D_{u}$ of the same ensemble member at timesteps $s<u$, and an intermediate timestep $t$, where $s < t < u$ (for FLINT*).  
The goal is to predict the corresponding flow field $\hat{F}_{t}$ and generate interpolated scalar fields $\hat{D}_{t}$ for any intermediate time $t\in [s,u]$.
To accomplish this, FLINT* first computes intermediate flow fields, $\Fts{}$ and $\Ftu{}$. The \emph{time-backward} flow field, $\Fts{}$, represents the flow vectors from the frame at time $t$ to an earlier frame at $s$. Conversely, the \emph{time-forward} flow, $\Ftu{}$, represents flow vectors from the frame at $t$  to a later frame at $u$. These intermediate flow fields are then used to warp scalar fields towards the target time $t$, generating estimates for that time step.
In the final step, \acronym combines the intermediate warped scalar fields using a fusion mask $M$ learned by FLINT*, where $M(i, j) \in [0, 1],\forall i, j$, which ensures smooth blending and high-quality interpolation.
\end{new}

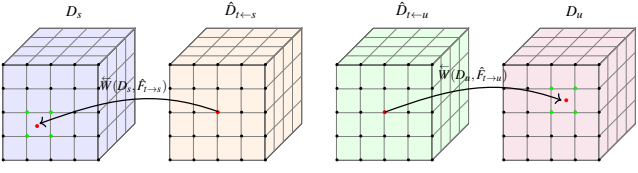
\begin{figure}[ht]
  \centering
  \resizebox{\linewidth}{!}{%
  \begin{tikzpicture}

    \draw[fill=blue!10] (0,0,2) -- (2,0,2) -- (2,2,2) -- (0,2,2) -- cycle;
    \draw[fill=blue!10] (2,0,0) -- (2,0,2) -- (2,2,2) -- (2,2,0) -- cycle;
    \draw[fill=blue!10] (0,2,0) -- (2,2,0) -- (2,2,2) -- (0,2,2) -- cycle;
    \node[label=above:$D_s$] at (1.5,2.75,2) {};

    \foreach \x in{0,0.5,...,2}
    {   \draw[gray, very thin] (0,\x ,2) -- (2,\x ,2);
        \draw[gray, very thin] (\x ,0,2) -- (\x ,2,2);
        \draw[gray, very thin] (2,\x ,2) -- (2,\x ,0);
        \draw[gray, very thin] (\x ,2,2) -- (\x ,2,0);
        \draw[gray, very thin] (2,0,\x ) -- (2,2,\x );
        \draw[gray, very thin] (0,2,\x ) -- (2,2,\x );
    }
    \foreach \x in {-0.76,-0.26,...,1.24}
        \foreach \y in {-0.76,-0.26,...,1.24}
            \filldraw (\x,\y,0) circle (0.6pt);
    \filldraw[red] (-0.05,-0.05,0) circle (0.85pt);
    \filldraw[green] (0.24,0.24,0) circle (0.75pt);
    \filldraw[green] (0.24,-0.26,0) circle (0.75pt);
    \filldraw[green] (-0.26,0.24,0) circle (0.75pt);
    \filldraw[green] (-0.26,-0.26,0) circle (0.75pt);

    \draw[fill=orange!10] (3.5,0,2) -- (5.5,0,2) -- (5.5,2,2) -- (3.5,2,2) -- cycle;
    \draw[fill=orange!10] (5.5,0,0) -- (5.5,0,2) -- (5.5,2,2) -- (5.5,2,0) -- cycle;
    \draw[fill=orange!10] (3.5,2,0) -- (5.5,2,0) -- (5.5,2,2) -- (3.5,2,2) -- cycle;
    \node[label=above:$\Dts{}$] at (5,2.75,2) {};

    \foreach \x in{3.5,4,4.5,5,5.5}
    {   
        \draw[gray, very thin] (3.5,\x-3.5,2) -- (5.5,\x-3.5,2);  
        \draw[gray, very thin] (\x ,0,2) -- (\x ,2,2);            
        \draw[gray, very thin] (5.5,\x-3.5,2) -- (5.5,\x-3.5,0);  
        \draw[gray, very thin] (\x ,2,2) -- (\x ,2,0);            
        \draw[gray, very thin] (5.5,0,\x-3.5) -- (5.5,2,\x-3.5);  
        \draw[gray, very thin] (3.5,2,\x-3.5) -- (5.5,2,\x-3.5);  
    }

    \foreach \x in {2.74,3.24,...,4.74}
        \foreach \y in {-0.76,-0.26,...,1.24}
            \filldraw (\x,\y,0) circle (0.6pt);
    \filldraw[red] (3.74,0.24,0) circle (0.85pt);

    \draw[fill=green!10] (7,0,2) -- (9,0,2) -- (9,2,2) -- (7,2,2) -- cycle;
    \draw[fill=green!10] (9,0,0) -- (9,0,2) -- (9,2,2) -- (9,2,0) -- cycle;
    \draw[fill=green!10] (7,2,0) -- (9,2,0) -- (9,2,2) -- (7,2,2) -- cycle;
    \node[label=above:$\Dtu{}$] at (8.6,2.75,2) {};

    \foreach \x in{7,7.5,8,8.5,9}
    {   
        \draw[gray, very thin] (7,\x-7,2) -- (9,\x-7,2);  
        \draw[gray, very thin] (\x ,0,2) -- (\x ,2,2);    
        \draw[gray, very thin] (9,\x-7,2) -- (9,\x-7,0);  
        \draw[gray, very thin] (\x ,2,2) -- (\x ,2,0);    
        \draw[gray, very thin] (9,0,\x-7) -- (9,2,\x-7);  
        \draw[gray, very thin] (7,2,\x-7) -- (9,2,\x-7);  
    }
    \foreach \x in {6.24,6.74,...,8.24}
        \foreach \y in {-0.76,-0.26,...,1.24}
            \filldraw (\x,\y,0) circle (0.6pt);
    \filldraw[red] (7.24,0.24,0) circle (0.85pt);

    \draw[fill=purple!10] (10.5,0,2) -- (12.5,0,2) -- (12.5,2,2) -- (10.5,2,2) -- cycle;
    \draw[fill=purple!10] (12.5,0,0) -- (12.5,0,2) -- (12.5,2,2) -- (12.5,2,0) -- cycle;
    \draw[fill=purple!10] (10.5,2,0) -- (12.5,2,0) -- (12.5,2,2) -- (10.5,2,2) -- cycle;
    \node[label=above:$D_u$] at (12,2.75,2) {};

    \foreach \x in{10.5,11,11.5,12,12.5}
    {   
        \draw[gray, very thin] (10.5,\x-10.5,2) -- (12.5,\x-10.5,2);  
        \draw[gray, very thin] (\x ,0,2) -- (\x ,2,2);                
        \draw[gray, very thin] (12.5,\x-10.5,2) -- (12.5,\x-10.5,0);  
        \draw[gray, very thin] (\x ,2,2) -- (\x ,2,0);                
        \draw[gray, very thin] (12.5,0,\x-10.5) -- (12.5,2,\x-10.5);  
        \draw[gray, very thin] (10.5,2,\x-10.5) -- (12.5,2,\x-10.5);  
    }

    \foreach \x in {9.74,10.24,...,11.99}
        \foreach \y in {-0.76,-0.26,...,1.24}
            \filldraw (\x,\y,0) circle (0.6pt);
    \filldraw[red] (11.05,0.49,0) circle (0.85pt);
    \filldraw[green] (10.74,0.24,0) circle (0.75pt);
    \filldraw[green] (11.24,0.24,0) circle (0.75pt);
    \filldraw[green] (11.24,0.74,0) circle (0.75pt);
    \filldraw[green] (10.74,0.74,0) circle (0.75pt);

    \draw[->, line width=0.75pt] (3.75,0.25,0) to[out=160,in=20] node[above=0.01cm, font=\footnotesize, xshift=3pt] {$\W ( D_s, \hat{F}_{t \rightarrow s})$} (0,0,0);
    \draw[->, line width=0.75pt] (7.24,0.24,0) to[out=20,in=160] node[above=0.01cm, font=\footnotesize] {$\W ( D_{u}, \hat{F}_{t \rightarrow u})$} (10.95,0.49,0);

  \end{tikzpicture}
  }
    \caption{ Illustration of the 3D backward warping $\W$:
    (scalar) fields $D_s$ and $D_u$ are reversely mapped according to
    the flow fields $\hat{F}_{t \rightarrow s}$ and $\hat{F}_{t \rightarrow u}$. 
    The fields $\Dts{}$ and $\Dtu{}$ are then reconstructed using trilinear interpolation
    considering the values at the coordinates shown with green dots (for the visible front surface of the cube). 
    }
  \label{fig:backward_warp_3d}
  \vspace{-6pt}
\end{figure}

\vspace{-3pt}
\textbf{Warping}.
In \acronym, we implement the volumetric \textit{backward warping}, where each target voxel \( v_t \) identifies its corresponding source voxel \( v_s \) based on the flow fields.
This mapping, guided by the intermediate flow fields, ensures smooth resampling using trilinear interpolation around \( v_s \) to compute the value for \( v_t \).
We denote the combined effect of reverse mapping and interpolation by the warping operator \( \W \).
Specifically, the mappings are guided by the flow fields, producing the warped scalar fields \(\Dts{} = \W (D_s, \Fts{})\) and \(\Dtu{} = \W (D_u, \Ftu{})\), which represent the values at time \( t \) based on the source volumes \( D_s \) and \( D_u \), respectively, see~\autoref{fig:backward_warp_3d}. 
 

\vspace{-6pt}

In \acronym, iterative refinement within FLINT*’s convolutional blocks is applied similarly to the original FLINT method~\cite{gadirov2024flint}. However, FLINT* is optimized specifically for use with a hypernetwork (see \autoref{subsec:hyper_vs_flint}), the architecture of which was constructed based on a thorough hyperparameter search (\autoref{subsec:param_study}) to ensure efficiency and compatibility.
The interpolated scalar field \(\hat{D}_t\), intermediate flow fields \(\Ft{i}\), and estimated final flow field \(\Fhat_t\) are computed as follows:
\vspace{-2pt}
\begin{new}
\begin{subequations}
\scalefont{0.99}
\begin{align}
    \hat{D}_t &= \Dts{N-1} \odot M + \Dtu{N-1} \odot (\mathbf{I} - M)\label{eq:interpolD}, \\
    \Ft{i+1} &= \Ftu{i}, \hspace{2pt} \text{for } i = 0, \dots, N-2, \quad\hat{F}_t =  \Ft{N-1} \hspace{2pt}(N=3).
    \label{eq:interpolF}
\end{align}
\end{subequations}
\end{new}

\vspace{-15pt}
\textbf{Inference}.
During the inference phase, \acronym processes the input scalar fields $D_s$ and $D_{u}$, along with their associated simulation parameters using the fully trained FLINT* and HyperNet networks. Here, GT information and loss functions are absent, as the network operates solely with its learned parameters. The hypernetwork plays a crucial role by dynamically generating the convolutional layer weights (i.e., kernels) for the FLINT* network based on the provided simulation parameters, as illustrated in~\autoref{fig:model}. This allows FLINT* to adapt its operations according to the specific characteristics of each input subset.
The final outputs, namely the interpolated scalar field $\Dhat_{t}$ and the reconstructed flow field $\Fhat_t$, are computed according to Eqs.~\eqref{eq:interpolD} and \eqref{eq:interpolF} respectively, see~\autoref{fig:nn_overview}.

\vspace{-3pt}
\subsection{Loss Function}
\vspace{-2pt}
\label{subsec:loss}
    
The total \acronym loss is a linear combination of reconstruction loss $\mathcal{L}_{rec}$ and flow loss $\mathcal{L}_{flow}$:
\begin{equation}\label{eq:loss-with-flow}
  \begin{aligned}
    \mathcal{L} = \mathcal{L}_{rec} + \lambda_{flow} \,\mathcal{L}_{flow},
  \end{aligned}
\end{equation}
where $\lambda_{flow} = 0.2$ for balancing total loss scale w.r.t.\
the reconstruction component (determined experimentally, see \autoref{subsec:param_study}).

\vspace{-3pt}
\textbf{Scalar field interpolation.}
To interpolate scalar fields temporally, we include a loss component to improve the accuracy of the interpolated density field from~Eq.\eqref{eq:interpolD}. The reconstruction loss $\Lrec$ measures the $L_1$ distance between the GT density $\DGT$ and the interpolated field $\hat{D}_{t}$ produced by the \acronym network: 
\begin{equation} 
\label{eq:Lrec}
\begin{aligned} 
\mathcal{L}_{rec} = \lVert \DGT - \hat{D}_{t} \rVert_{1}. 
\end{aligned} 
\end{equation} 
This ensures that the interpolated scalar field closely matches the GT, improving the accuracy of the temporal interpolation.

\vspace{-3pt}
\textbf{Flow estimation}.
To improve the quality of the learned flow field, we incorporate a flow loss component calculated as the $L_1$ distance between the estimated flow at each network block and the GT flow (used only during training). Accumulating this measure across all blocks, rather than only the final one, yields better results. Additionally, we adopt exponentially increasing weights for the loss from RAFT~\cite{teed2020raft}, resulting in the following flow loss equation:
\begin{equation}
  \mathcal{L}_{\text{flow}} = 
  \sum_{i=1}^{N} \gamma^{N-i} \lVert \FGT - \hat{F}^{i}_{t} \rVert_{1},
\end{equation}
where $\FGT$ is the GT flow at time $t$,
$\hat{F}^{i}_{t}$ is the flow output from the
corresponding $i^{th}$ block of the FLINT*
network (Eq.~\eqref{eq:interpolF}), and $N=3$ is the number of blocks in the model.
We experimentally established the value of $\gamma$
as 0.8, aligning with the RAFT loss and validating
this choice through hyperparameter search.

\vspace{-3pt}
\subsection{Comparison between \acronym and FLINT methods}
\vspace{-2pt}
\label{subsec:hyper_vs_flint}
The key innovation of \acronym is its integration of a hypernetwork that dynamically generates weights for the FLINT* network, a variant of the original FLINT model where the student-teacher setup has been removed, thus reducing complexity while being optimized to work with the hypernetwork. This enables \acronym to condition convolutional layers on simulation parameters, aligning the network more precisely with data characteristics. The result is enhanced accuracy in flow estimation and scalar field interpolation, capturing physical phenomena with greater fidelity than the original FLINT approach.
Furthermore, the inclusion of a hypernetwork introduces a novel capability to \acronym: parameter space exploration. By conditioning the model on input parameters, \acronym generates predictions for unseen configurations, enabling researchers to estimate outcomes for unsimulated parameter sets. This approach is particularly valuable in computationally expensive scientific studies, offering insights without requiring exhaustive simulations. This flexibility distinguishes \acronym from traditional methods like FLINT, which lack parameter-driven adaptability.



\section{Study Setup}
\vspace{-2pt}
\label{sec:study_setup}
	
In this section, we describe the training setup, provide an overview of the datasets used in our experiments and discuss the evaluation methods employed to assess the results obtained by \acronym.

\vspace{-3pt}
\subsection{Training}
\vspace{-2pt}
\label{subsec:training}
We apply the standard prepossessing step of normalization to [0, 1] for the scalar fields and to [-1, 1] for the vector fields before the start of the training.
\acronym is optimized using AdamW~\cite{loshchilov2018fixing} with
early stopping, similarly to \cite{gadirov2024flint} to prevent overfitting on the training data.
We use an experimentally determined learning rate of $10^{-4}$ with a cosine annealing scheduler that gradually decreases the learning rate
to $10^{-5}$ by the end of the training.
We train \acronym with mini-batches of size 4 for the 3D ensemble datasets used in our study (see~\autoref{subsec:data}).
We split the set of all available data into training,
validation, and test subsets.
To support arbitrary interpolation and flow estimation during training, $t \in [s, u]$ is chosen randomly within a maximum time window of size 12, similarly to \cite{gadirov2024flint}. This window, confirmed through hyperparameter search, efficiently determines the maximum time gap between sampled timesteps $s$ and $u$ for constructing the training set, allowing for effective interpolation.
Our proposed \acronym model is trained for 200 epochs for 8 hours on a single Nvidia Titan V GPU with 12GB of VRAM to converge for all datasets.

\vspace{-3pt}
\subsection{Datasets and Evaluation}
\vspace{-2pt}
\label{subsec:data}
We consider two scientific ensemble datasets in our study.

\vspace{-3pt}
\textbf{Nyx.}
The first dataset is a 3D+time ensemble based on the compressible cosmological hydrodynamics simulation Nyx, developed by Lawrence Berkeley National Laboratory~\cite{sexton2021nyx}.
We consider an ensemble comprising 36 members, each consisting of a maximum of 1600 timesteps with a spatial resolution of $128\times 128\times 128$.
It contains density and the $x$, $y$, $z$ components of velocity.
Akin to InSituNet~\cite{he2019insitunet}, we vary three parameters for ensemble generation: the total matter density ($\Omega_m \in [0.1, 0.2]$), the total density of baryons ($\Omega_b \in [0.0215, 0.0235]$), and the Hubble constant ($h \in [0.55, 0.75]$).
We randomly sample a training subset of 500 timesteps and utilize different ensemble members for training, validation, and testing.

\vspace{-3pt}
\textbf{Castro.}
The second dataset consists of a 3D+time ensemble based on the astrophysical hydrodynamics simulation Castro, simulating the merger of two white dwarfs, developed by the Lawrence Berkeley National Laboratory~\cite{2010castro}. 
This ensemble contains 12 members, each with up to 800 timesteps, and a spatial resolution of $128 \times 128 \times 128$. 
The dataset includes the density field and the $x$, $y$, and $z$ velocity components. 
For ensemble generation, we vary two parameters based on the authors’ suggestions, such as the masses of the primary ($M_P$) and secondary ($M_S$) white dwarfs ($M_P, M_S \in [0.8, 0.95] \, M_\odot$), where $M_\odot$ represents the solar mass.
A training subset of 400 timesteps is randomly sampled, with different ensemble members designated for training, validation, and testing.


\vspace{-3pt}
We evaluate \acronym's performance in reconstructing scalar and vector fields both qualitatively and quantitatively. For density field evaluation, we use \textit{peak signal-to-noise ratio} (PSNR), while the accuracy of the flow field is assessed with \textit{endpoint error} (EPE), which calculates the average Euclidean distance between estimated and GT flow vectors—lower EPE values indicating higher accuracy. For qualitative assessment, we visualize flow field outcomes for the simulation ensembles. Given the 3D nature of our datasets, PSNR and EPE are calculated in the volume domain, ensuring the metrics align with the spatial characteristics of the data and provide a robust evaluation of \acronym's performance.


\begin{figure*}[!t]
    \centering
    \scalebox{0.92}{
    \begin{minipage}[t]{0.02\linewidth}
        \vspace*{-495pt}
        \text{\rotatebox{90}{\footnotesize Dens GT}} \\ [25pt]
        \text{\rotatebox{90}{\footnotesize Dens \acronym}} \\ [20pt]
        \text{\rotatebox{90}{\footnotesize Dens FLINT}} \\ [30pt]
        \text{\rotatebox{90}{\footnotesize Dens STSR-INR}} \\ [28pt]
        \text{\rotatebox{90}{\footnotesize Flow GT}} \\ [28pt]
        \text{\rotatebox{90}{\footnotesize Flow \acronym}} \\ [20pt]
        \text{\rotatebox{90}{\footnotesize Flow FLINT}} \\
    \end{minipage}
    \hspace*{1pt}
    \begin{minipage}[t]{0.48\linewidth}
        \begin{tikzpicture}
            \node (mainfig) at (0,0) {
                \includegraphics[trim=640 63 606 63, clip, width=0.93\linewidth]{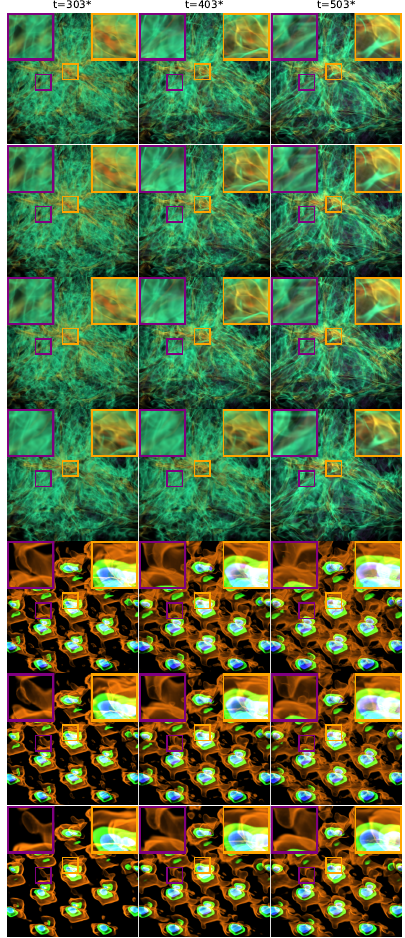}
            };
        \end{tikzpicture}
        \vspace{-5pt}
        \subcaption{Nyx results}
        \label{fig:nyx_results}
    \end{minipage}
        \begin{minipage}[t]{0.48\linewidth}
        \begin{tikzpicture}
            \node (mainfig) at (0,0) {
                \includegraphics[trim=640 63 606 63, clip, width=0.93\linewidth]{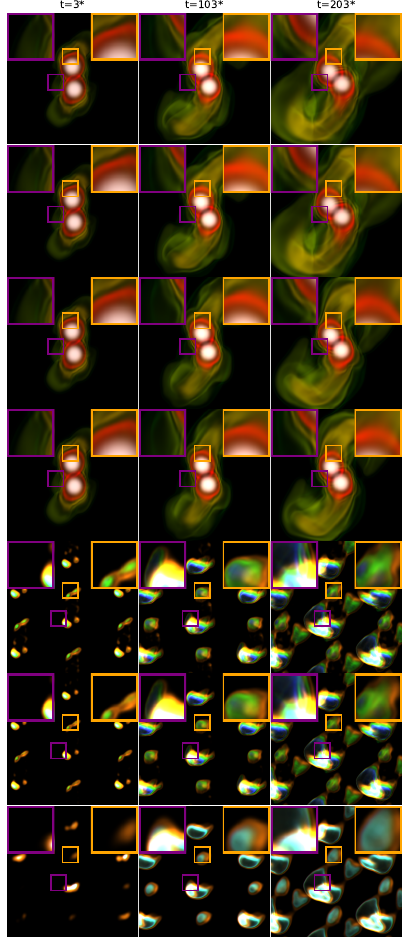}
            };
        \end{tikzpicture}
        \vspace{-5pt}
        \subcaption{Castro results}
        \label{fig:castro_results}
    \end{minipage}
    }
    \vspace{-5pt}
    \caption{Nyx and Castro: \acronym flow field estimation and temporal density interpolation, 5$\times$.
        From top to bottom, the rows show GT density, \acronym interpolated density, FLINT interpolation, STSR-INR interpolation, GT flow, \acronym flow estimation, and FLINT flow estimation. 
        3D rendering was used for the density and flow visualization (\protect\orangecircle{} \protect\greencircle{} \protect\bluecircle{} colors representing $x$, $y$, and $z$ flow directions respectively).
        \vspace{-18pt}
        }
    \label{fig:nyx_castro}
\end{figure*}

\section{Qualitative Results}
\vspace{-2pt}
\label{sec:results}
We evaluate \acronym on the two datasets (\autoref{subsec:data}) with respect to flow estimation and density interpolation (i.e., temporal super-resolution), and compare it with FLINT and STSR-INR.

\vspace{-3pt}
\subsection{Nyx}
\vspace{-2pt}

\label{subsec:physical_flow_nyx}
First, we consider the scenario where GT density and velocity fields are available for some members of the entire 3D ensemble.
As \autoref{fig:nyx_results} illustrates, \acronym achieves accurate performance in terms of both density field interpolation and flow field estimation.
Visually, the difference between the renderings of the reconstructed density field (second row) and its GT is minimal. 
Moreover, even at a relatively high interpolation rate of $5\times$, \acronym effectively learns a flow field that structurally resembles the GT flow.
Both \acronym and FLINT (third row) produce results that are visually very close to the GT for density interpolation; however, when compared to STSR-INR (fourth row), \acronym demonstrates more accurate density field interpolation.
At $t=303$, $t=403$, and $t=503$, the purple and orange zoom-ins clearly reveal a different structure and less dark matter density in STSR-INR's results compared to \acronym, which reconstructs density more effectively.

\vspace{-3pt}
Moreover, \acronym not only reconstructs the density field but also supplements it with accurate flow information, as shown in \autoref{fig:nyx_results} (sixth row), a feature that STSR-INR lacks.
When examining the flow, we observe circular swirling patterns, indicating the complex dynamics of the baryonic gas. 
As these flows intensify, we see evidence of dark matter moving outward---reflected in both the GT and \acronym density, especially in the orange zoom-ins---consistent with an expanding universe.
This underlines the utility of \acronym in capturing not just the static density fields but also the dynamic evolution of the cosmic structures.
Furthermore, \acronym outperforms the FLINT model (last row) in capturing flow information, producing more accurate representations of the underlying dynamics. 
For example, at $t=303$ and $t=503$, in the purple and orange zoom-ins, the FLINT flow exhibits structural differences compared to the GT, whereas \acronym more effectively preserves the intricate flow dynamics.

\begin{new}
\vspace{-3pt}
A domain expert from astronomy specializing in cosmological simulations and observational data highlighted the advantages of estimating both density and velocity fields. This capability is crucial in cosmology for estimating distances between astronomical objects and analyzing their spatial relationships. This would help in bridging simulations with real-world observations.  Velocity estimation is particularly valuable for constructing ``lightcones'', in which simulation data are used to model the evolution of the universe.

\end{new}

\vspace{-3pt}
\subsection{Castro}
\vspace{-2pt}
For the Castro ensemble dataset, as \autoref{fig:castro_results} illustrates, \acronym achieves  results that are visually very close to the GT for both density field interpolation and flow field estimation.
When comparing \acronym to FLINT and STSR-INR for temporal density interpolation (third and fourth row), \acronym demonstrates superior accuracy. For instance, at $t=103$ and $t=203$, STSR-INR and FLINT show noticeable deviations in structure and reduced matter density around the merging white dwarfs. In contrast, \acronym preserves these density structures more effectively, aligning well with the GT.
\begin{new}
For flow estimation, \acronym produces more coherent and spatially accurate flow fields than FLINT. This is particularly evident at $t=3$ and $t=203$ in \autoref{fig:castro_results}, where FLINT introduces artifacts and inconsistencies in motion direction. In contrast, \acronym better captures the developing flow dynamics, especially in the detailed areas in the purple and orange zoom-ins, where it maintains a more accurate representation of the evolving flows.
\end{new}





\section{Quantitative and Comparative Evaluation}
\vspace{-2pt}
\label{sec:compar_eval}
	
In this section, we present quantitative results and compare against
baseline methods to demonstrate the improvement achieved with our
proposed method, followed by ablation and parameter studies to explore different configurations and hyperparameters.

\vspace{-3pt}
\subsection{Comparison Against Baselines}
\vspace{-2pt}


STSR-INR~\cite{tang2024stsr} and CoordNet~\cite{han2022coordnet} were evaluated as key benchmarks for temporal super-resolution (TSR) tasks on the 3D+time Nyx and Castro datasets, with results shown in \autoref{nyx_comparison} and \autoref{castro_comparison}. STSR-INR employs a variational auto-decoder to optimize latent vectors for variable interpolation in the latent space, building on prior works like STNet~\cite{han2021stnet}.
It has demonstrated notable improvements in density interpolation but focuses solely on scalar field interpolation, lacking capabilities for flow field estimation.
Similarly, CoordNet, a benchmark for visualizing time-varying volumetric data, also improves upon TSR-TVD~\cite{han2019tsr} as detailed in \autoref{sec:related_work}. While CoordNet achieves strong results in PSNR scores for density interpolation across various rates, it shares STSR-INR’s limitation in addressing only TSR of scalar fields without extending to flow field estimation.
In contrast, \acronym not only outperforms both methods in density interpolation but also introduces flow estimation, a capability absent in both CoordNet and STSR-INR. 

\vspace{-4pt}
FLINT~\cite{gadirov2024flint}, based on a pure CNN architecture with a student-teacher training mechanism, serves as another valuable benchmark for comparison. FLINT demonstrated improved results over STSR-INR and CoordNet in temporal interpolation and provided reasonable flow estimation, making it a strong competitor. Our evaluation against FLINT shows that \acronym slightly outperforms FLINT in terms of TSR for scalar fields. 
More notably, \acronym demonstrates superior performance in flow estimation, yielding significantly better results. This is evidenced by the lower EPE scores reported for both the Nyx and Castro datasets, as shown in \autoref{nyx_comparison} and \autoref{castro_comparison}.
Furthermore, \acronym offers the added advantage of enabling parameter space exploration, a feature neither FLINT nor other baselines can achieve. This makes \acronym a more versatile and powerful tool for analyzing scientific ensembles.

\vspace{-6pt}
\begin{table}[h]
\scalefont{0.78}
\centering
  \caption{ Comparison against baselines, Nyx }
  \vspace{-8pt}
  \begin{tabular}{lc@{\hspace{5pt}}c@{\hspace{3pt}}c|c@{\hspace{5pt}}c@{\hspace{3pt}}c|c@{\hspace{5pt}}c@{\hspace{3pt}}c} 
    \toprule
    \multirow{2}{*}{\textbf{Method}} & \multicolumn{3}{c}{\textbf{3$\times$}} & \multicolumn{3}{c}{\textbf{5$\times$}} & \multicolumn{3}{c}{\textbf{8$\times$}} \\
    \cmidrule(lr){2-4} \cmidrule(lr){5-7} \cmidrule(lr){8-10}
    & \textbf{PSNR $\uparrow$} & \textbf{EPE $\downarrow$} & & \textbf{PSNR $\uparrow$} & \textbf{EPE $\downarrow$} & & \textbf{PSNR $\uparrow$} & \textbf{EPE $\downarrow$} & \\
    \midrule
    \acronym & \textbf{53.32} & \textbf{0.0237} & & \textbf{52.70} & \textbf{0.0238} & & \textbf{51.07} & \textbf{0.0242} & \\
    FLINT & 53.17 & 0.0310 & & 52.31 & 0.0310 & & 49.39 & 0.0309 & \\
    STSR-INR & 49.63 & --- & & 44.21 & --- & & 41.09 & --- & \\
    CoordNet & 49.37 & --- & & 44.02 & --- & & 40.78 & --- & \\
    Linear & 47.49 & --- & & 41.92 & --- & & 37.51 & --- & \\
    \bottomrule
  \end{tabular}
  \label{nyx_comparison}
\vspace{-3pt}
\end{table}

\begin{table}[h]
\scalefont{0.78}
\centering
  \caption{ Comparison against baselines, Castro }
  \vspace{-6pt}
  \begin{tabular}{lc@{\hspace{5pt}}c@{\hspace{3pt}}c|c@{\hspace{5pt}}c@{\hspace{3pt}}c|c@{\hspace{5pt}}c@{\hspace{3pt}}c}
    \toprule
    \multirow{2}{*}{\textbf{Method}} & \multicolumn{3}{c}{\textbf{3$\times$}} & \multicolumn{3}{c}{\textbf{5$\times$}} & \multicolumn{3}{c}{\textbf{8$\times$}} \\
    \cmidrule(lr){2-4} \cmidrule(lr){5-7} \cmidrule(lr){8-10}
    & \textbf{PSNR $\uparrow$} & \textbf{EPE $\downarrow$} & & \textbf{PSNR $\uparrow$} & \textbf{EPE $\downarrow$} & & \textbf{PSNR $\uparrow$} & \textbf{EPE $\downarrow$} & \\
    \midrule
    \acronym & \textbf{49.48} & \textbf{0.0275} & & \textbf{47.39} & \textbf{0.0276} & & \textbf{45.41} & \textbf{0.0278} & \\
    FLINT & 47.89 & 0.0502 & & 46.16 & 0.0506 & & 43.83 & 0.0513 & \\
    STSR-INR & 44.83 & --- & & 42.38 & --- & & 40.26 & --- & \\
    CoordNet & 44.52 & --- & & 42.29 & --- & & 40.08 & --- & \\
    Linear & 42.11 & --- & & 37.28 & --- & & 33.64 & --- & \\
    \bottomrule
  \end{tabular}
  \label{castro_comparison}
\vspace{-6pt}
\end{table}

We conducted an inference time comparison of \acronym against FLINT, CoordNet, and STSR-INR across the two 3D ensemble datasets, measuring inference time over 300 timesteps from the test set. \acronym significantly outperforms CoordNet and STSR-INR in speed, achieving an average of 0.18 seconds per timestep compared to 2.1 seconds for CoordNet and 1.5 seconds for STSR-INR. Remarkably, \acronym even slightly outpaces FLINT, which averages 0.2 seconds per timestep. This improvement arises from \acronym's efficient CNN utilization, which, despite the added hypernetwork complexity, maintains performance advantage. Unlike INR-based methods that solve implicit functions at each point, \acronym leverages CNNs' parallel processing capabilities and optimized memory access, resulting in faster inference.


\vspace{-10pt}
\subsection{Ablation Studies}
\vspace{-2pt}
    \label{subsect:ablation}
    Our proposed \acronym method is a deep neural network that has
    various loss components.
    To assess the impact of these
    in achieving optimal results, we conducted a series of ablation
    studies.

\begin{table}[h]
\vspace{-6pt}
\scalefont{0.8}
  \centering
  \caption{Ablation of \acronym}
  \vspace{-6pt}
  \begin{tabular}{lc@{\hspace{5pt}}c@{\hspace{3pt}}c|c@{\hspace{5pt}}c@{\hspace{3pt}}c@{\hspace{5pt}}c}
    \toprule
    \textbf{Method} & \multicolumn{3}{c}{\textbf{Nyx, 5$\times$}} & \multicolumn{3}{c}{\textbf{Castro, 5$\times$}} \\
    \cmidrule(lr){2-4} \cmidrule(lr){5-7} 
    & \textbf{PSNR $\uparrow$} & \textbf{EPE $\downarrow$} & & \textbf{PSNR $\uparrow$} & \textbf{EPE $\downarrow$} & \\
    \midrule
    \acronym no \textit{flow} & 51.94 & 0.1348 & & 46.92 & 0.7432 & \\
    \acronym no \textit{rec} & 44.78 & 0.0254 & & 37.89 & 0.0335 & \\
    \acronym w/o hyper & 50.89 & 0.0357 & & 46.04 & 0.0516 & \\
    \acronym & \textbf{52.70} & \textbf{0.0238} & & \textbf{47.39} & \textbf{0.0276} & \\
    \bottomrule
  \end{tabular}
  \label{ablation}
 \vspace{-6pt}
\end{table}

The ablation studies on \acronym reveal the impact of various loss components on performance, as shown in~\autoref{ablation} for the Nyx and Castro datasets. Four variants are compared:  
\begin{itemize}
    \item \acronym no \textit{flow}: omits the flow loss, leading to poor flow estimation (EPE) despite decent density interpolation (PSNR).
    \item \acronym no \textit{rec}: omits the reconstruction loss, resulting in decent flow learning but compromised density interpolation.  
    \item \acronym w/o hyper: removes the hypernetwork, leading to subpar interpolation and flow estimation due to the absence of dynamic adaptation enabled by simulation parameters.  
    \item \acronym: the full model with all components, achieves the highest scores in both PSNR and EPE.
\end{itemize}


\vspace{-10pt}
\subsection{Hyperparameter Studies}
\vspace{-2pt}
\label{subsec:param_study}
	
Our proposed \acronym method involves several hyperparameters, and we conducted extensive hyperparameter optimization to identify the optimal configuration for \acronym, testing variations in model width, depth, and loss functions.
\autoref{hyperparameter_search} displays the outcomes of these studies regarding spatio-temporal datasets.

\setlength{\tabcolsep}{2pt} 
\renewcommand{\arraystretch}{0.8} 

\vspace{-6pt}
\begin{table}[ht]
\centering
\caption{Hyperparameter search for \acronym}
\vspace{-6pt}
\scalefont{0.75}
\begin{tabular}{l@{\hspace{10pt}}c@{\hspace{10pt}}c|c@{\hspace{10pt}}c@{\hspace{10pt}}c}
\toprule
\textbf{Method} & \multicolumn{2}{c}{\textbf{Nyx, 5$\times$}} & \multicolumn{2}{c}{\textbf{Castro, 5$\times$}} \\
\cmidrule(lr){2-3} \cmidrule(lr){4-5}
& \textbf{PSNR $\uparrow$} & \textbf{EPE $\downarrow$} & \textbf{PSNR $\uparrow$} & \textbf{EPE $\downarrow$} \\
\midrule
\acronym\textit{64} & 51.68 & 0.0315 & 47.12 & 0.0430 & \\
\acronym\textit{Lapl} & 51.73 & 0.0319 & 44.95 & 0.0480 \\
\acronym\textit{hyper all} & 48.19 & 0.0281 & 42.79 & 0.0419 \\
\acronym\textit{w teacher} & 49.31 & 0.0265 & 44.43 & 0.0393 \\
\acronym\textit{$\lambda_{flow} = 0.3$} & 52.29 & 0.0311 & 45.39 & 0.0316 & \\
\acronym\textit{$\lambda_{flow} = 0.1$} & 52.19 & 0.0320 & 47.01 & 0.0325 & \\
\acronym\textit{$\gamma = 0.9$} & 52.14 & 0.0310 & 46.98 & 0.0345 & \\
\acronym\textit{$\gamma = 0.7$} & 52.28 & 0.0317 & 47.07 & 0.0371 \\
\acronym\textit{stride = 1} & 52.19 & 0.0418 & 46.94 & 0.0348 & \\
\acronym\textit{2 Blocks} & 49.72 & 0.0245 & 46.87 & 0.0407 \\
\acronym\textit{3 Blocks} & \textbf{52.70} & \textbf{0.0238} & \textbf{47.39} & \textbf{0.0276} \\
\acronym\textit{4 Blocks} & 52.51 & 0.0249 & 46.98 & 0.0314 & \\
\acronym\textit{5 Blocks} & 52.09 & 0.0256 & 46.97 & 0.0341 & \\
\acronym\textit{hyper no MLP} & 52.27 & 0.0261 & 46.06 & 0.0363 & \\
\acronym\textit{hyper no CNN} & 51.74 & 0.0282 & 45.46 & 0.0378 & \\
\acronym\textit{hyper no dropout} & 52.48 & 0.0259 & 46.61 & 0.0298 & \\
\bottomrule
\end{tabular}
\label{hyperparameter_search}
\vspace{-6pt}
\end{table}

The best-performing setup, labeled ``\acronym\textit{3 Blocks}'', includes three blocks with convolutional layers having channel counts decreasing from 128 to 64, balancing capacity and avoiding overfitting. Alternatives like ``\acronym\textit{hyper all}'' (all layers affected by the hypernetwork) and ``\acronym\textit{w teacher}'' (using a teacher block similar to FLINT) resulted in convergence issues and degraded performance. Similarly, the ``\acronym\textit{stride = 1}'' variant underperformed due to reduced expressiveness from fixed strides.
We explored loss functions, comparing ``\acronym\textit{Lapl}'' using a Laplacian pyramid, and ``\acronym\textit{3 Blocks}'' which applies $L_1$ loss. The simpler $L_1$ loss proved more effective. Additionally, we show the roles of HyperNet components, as configurations like ``\acronym\textit{hyper no MLP},'' ``\acronym\textit{hyper no CNN},'' and ``\acronym\textit{hyper no dropout}'' exhibited reduced performance. These findings underscore the importance of the proposed architecture and individual components.
\begin{new}
In~\autoref{hyperparameter_search_64}, we present the hyperparameter search results for Nyx at $64^3$ resolution, where ``\acronym\textit{3 Blocks}'' remains the best-performing model, consistent with what has been determined in~\autoref{hyperparameter_search} at $128^3$. This suggests good generalization across resolutions, with a slight performance drop at higher resolutions due to increased data complexity.
\end{new}

\begin{table}[ht]
\centering
\vspace{-6pt}
\scalefont{0.8}
\caption{\begin{new} Hyperparameter search for \acronym, Nyx, 5$\times$, $64^3$\end{new}}
\vspace{-6pt}
\begin{tabular}{l@{\hspace{10pt}}c@{\hspace{10pt}}c}
\toprule
\textbf{Method} & \textbf{PSNR $\uparrow$} & \textbf{EPE $\downarrow$} \\
\midrule

\begin{new}\acronym\textit{hyper all}\end{new} & 48.87 & 0.0279 \\
\begin{new}\acronym\textit{3 Blocks}\end{new} & \textbf{52.93} & \textbf{0.0220} \\
\begin{new}\acronym\textit{4 Blocks}\end{new} & 52.81 & 0.0231 \\
\begin{new}\acronym\textit{hyper no MLP}\end{new} & 52.33 & 0.0254 \\
\begin{new}\acronym\textit{hyper no CNN}\end{new} & 51.65 & 0.0277 \\
\bottomrule
\end{tabular}
\label{hyperparameter_search_64}
\vspace{-6pt}
\end{table}

\vspace{-5pt}
\section{Simulation Parameter Space Exploration}
\vspace{-2pt}
\label{sec:param_space_exp}

\begin{old}
Leveraging its HyperNet architecture, \acronym adapts to specific simulation parameters, enabling efficient analysis of parameter variations on simulation outcomes.
\end{old}
This section highlights two key examples demonstrating \acronym's utility in parameter space exploration. First, \autoref{subsec:similarity} examines the correlation between HyperNet-generated weights and simulation data, showcasing a strong alignment with simulation characteristics. Second, \autoref{subsec:param_data} demonstrates \acronym's ability to interpolate within the parameter space, accurately predicting outputs for configurations beyond the training set. These functionalities are validated using experiments with the Nyx simulation ensemble.


\begin{figure}[ht]
\vspace{-36pt}
    \centering
    \scalebox{0.9}{
    \begin{tikzpicture}
        \node[anchor=south west, inner sep=0] (image1) at (0, 0) {\includegraphics[width=\linewidth]{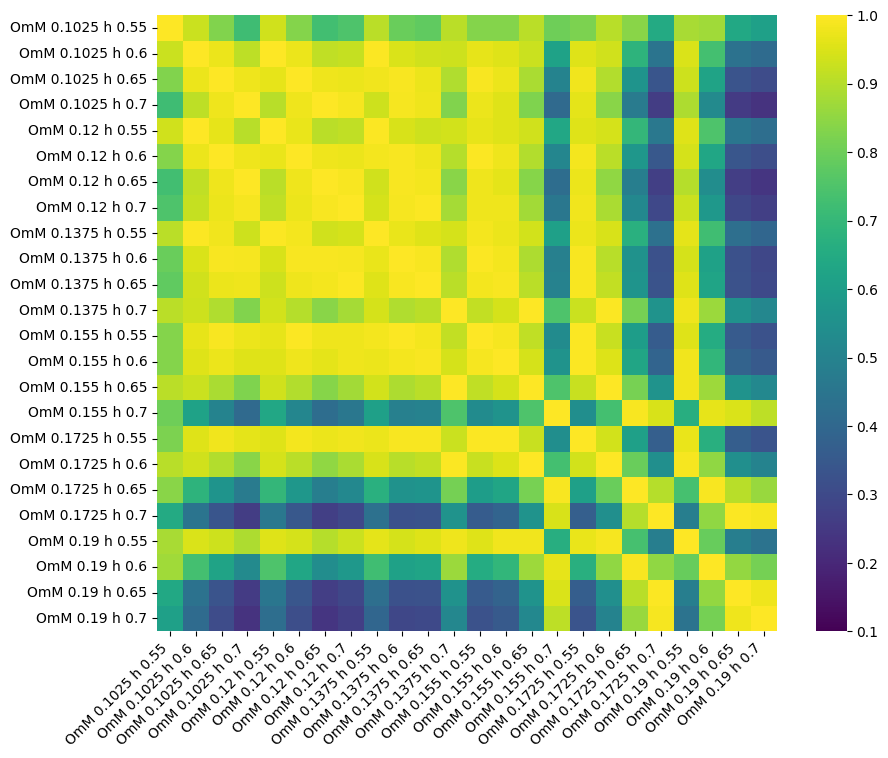}};
        \node[anchor=south west, inner sep=0] (image2) at (0, 0) {\includegraphics[width=\linewidth]{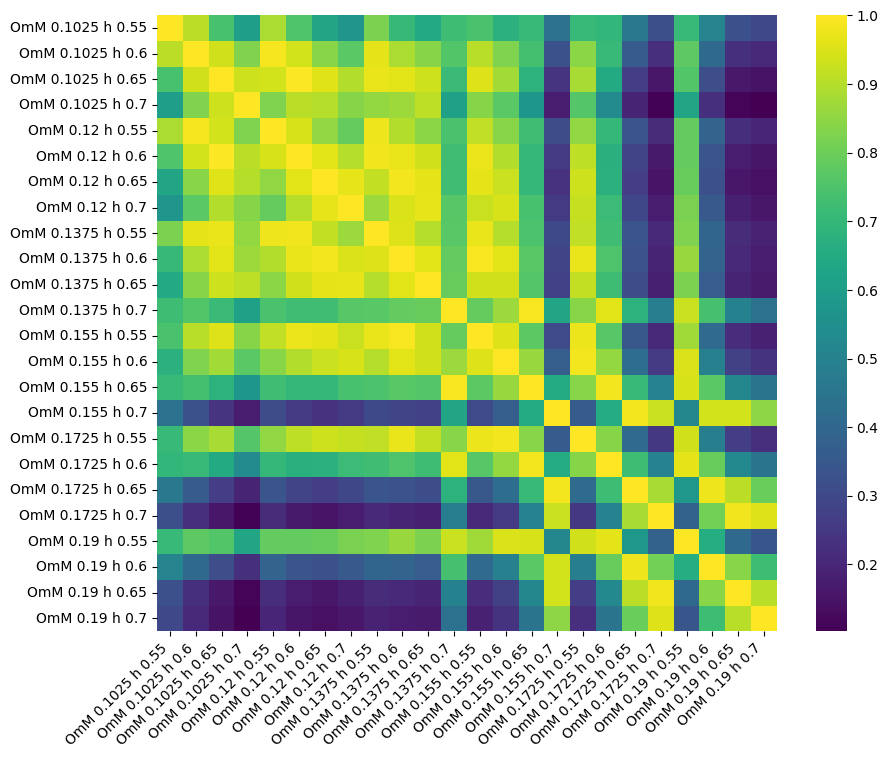}};
        
        \begin{scope}
            \clip
                (0,0) -- (0,8.6) -- (8.6,0) -- cycle; 
            \node[anchor=south west, inner sep=0] at (0,0) 
            {\includegraphics[width=\linewidth]{figures/Results/params/hypernet_similarity_matrix_OmM_h_24.png}};
        \end{scope}
        
        \begin{scope}
            \clip
                (0,8.6) -- (8.6, 8.6) -- (8.6,0) -- cycle; 
            \node[anchor=south west, inner sep=0] at (0,0) {\includegraphics[width=\linewidth]{figures/Results/params/mse_similarity_matrix_OmM_h_24.png}};
        \end{scope}
        
        \draw[thick, magenta] (1.3, 7.3) -- (7.6, 1.0);

    \node[magenta, above right] at (1.5, 7.0) {\small \textbf{Data}};
    \node[magenta, above left] at (1.3, 6.95) {\small \textbf{Weights}};
    \end{tikzpicture}
    }
    \vspace{-6pt}
    \caption{Similarity matrix: lower-left---HyperNet's weight similarity, upper-right---Nyx simulation parameter similarity. 
    }
    \label{fig:blended_similarity_matrix}
    \vspace{-18pt}
\end{figure}



\vspace{-10pt}
\subsection{Hypernetwork weights as Proxy for Data Similarity}
\vspace{-2pt}
\label{subsec:similarity}
To better understand the broader capabilities of HyperNet beyond generating weights for the FLINT* network, we explored its potential for parameter space analysis. Specifically, we constructed similarity matrices to examine how the weights generated by HyperNet correlate with the underlying data dynamics, reflecting the influence of simulation parameters. This analysis highlights how HyperNet weights, apart from serving as weight generation for the main network, can provide insights into parameter-driven variations within the dataset. 
Specifically, we compute similarity matrices for (i) the weights produced by HyperNet and (ii) the original dataset volumes (both in \autoref{fig:blended_similarity_matrix}), both of which are naturally influenced by the simulation parameters.
For this analysis, we considered 24 members from the Nyx ensemble under variation of
$\Omega_m = [0.1025, 0.12, 0.1375, 0.155, 0.1725, 0.19]$, 
$h = [0.55, 0.6, 0.65, 0.7]$,
while $\Omega_b$ was fixed at 0.0225.
Our proposed \acronym model was trained on 25\% of the data, specifically on six members with the combinations of parameters 
$\Omega_m = [0.1025, 0.1375, 0.19]$, $h = [0.55, 0.7]$, with $\Omega_b = 0.0225$.

\vspace{-3pt}
The similarity matrices reveal a strong correlation between the HyperNet-generated weights and the dataset volumes, confirming that the HyperNet effectively learns parameter-dependent representations. This correlation suggests that differences in HyperNet weights can serve as a meaningful surrogate for differences in the underlying data.
Upon closer examination, certain members stand out with distinguishable characteristics. For instance, for parameter configurations such as \(\Omega_m = 0.155\) and \(h = 0.7\), and \(\Omega_m = 0.1725\) with \(h = 0.65\) or \(h = 0.7\), the data shows lower similarity compared to other ensemble members. Similarly, for \(\Omega_m = 0.19\) with \(h = 0.6\), \(h = 0.65\), and \(h = 0.7\), the matrix highlights notable deviations in the data structure. 
Despite these variations, HyperNet’s weight similarities consistently maintain strong correlations across all configurations, enabling insights into data dynamics without requiring generation and access to the original data fields. 
We also conducted a quantitative evaluation using a \textit{triplet loss} approach~\cite{schultz2003learning} to validate the alignment between HyperNet-generated weights and parameter-dependent data dynamics. 
We analyzed triplets comprising an anchor, a more similar member, and a less similar member to evaluate how well distances in the HyperNet-generated weight space align with those in the data space. Specifically, we computed the distance between the anchor and both the more similar and less similar members in the data space and performed the same calculation in the HyperNet-generated weight space, confirming that weight similarity strongly correlates with data similarity.
Achieving a 96\% triplet correlation, this result underscores \acronym's utility in parameter space analysis, where learned representations guide tasks like constructing characteristic maps of parameter impacts and optimizing sampling strategies.





\begin{old}
\begin{figure*}[ht!]
    \centering
    \scalebox{0.92}{
    \begin{minipage}[t]{0.03\linewidth}
        \vspace*{-255pt}
        \text{\rotatebox{90}{\footnotesize Dens GT}} \\ [23pt]
        \text{\rotatebox{90}{\footnotesize Dens \acronym}} \\ [20pt]
        \text{\rotatebox{90}{\footnotesize Flow GT}} \\ [23pt]
        \text{\rotatebox{90}{\footnotesize Flow \acronym}} \\
    \end{minipage}
    \hspace*{3pt}
    \begin{minipage}[t]{0.82\linewidth}
        \begin{tikzpicture}
            \node (mainfig) at (0,0) {
                \includegraphics[trim=0 0 0 308, clip, width=0.98\linewidth]{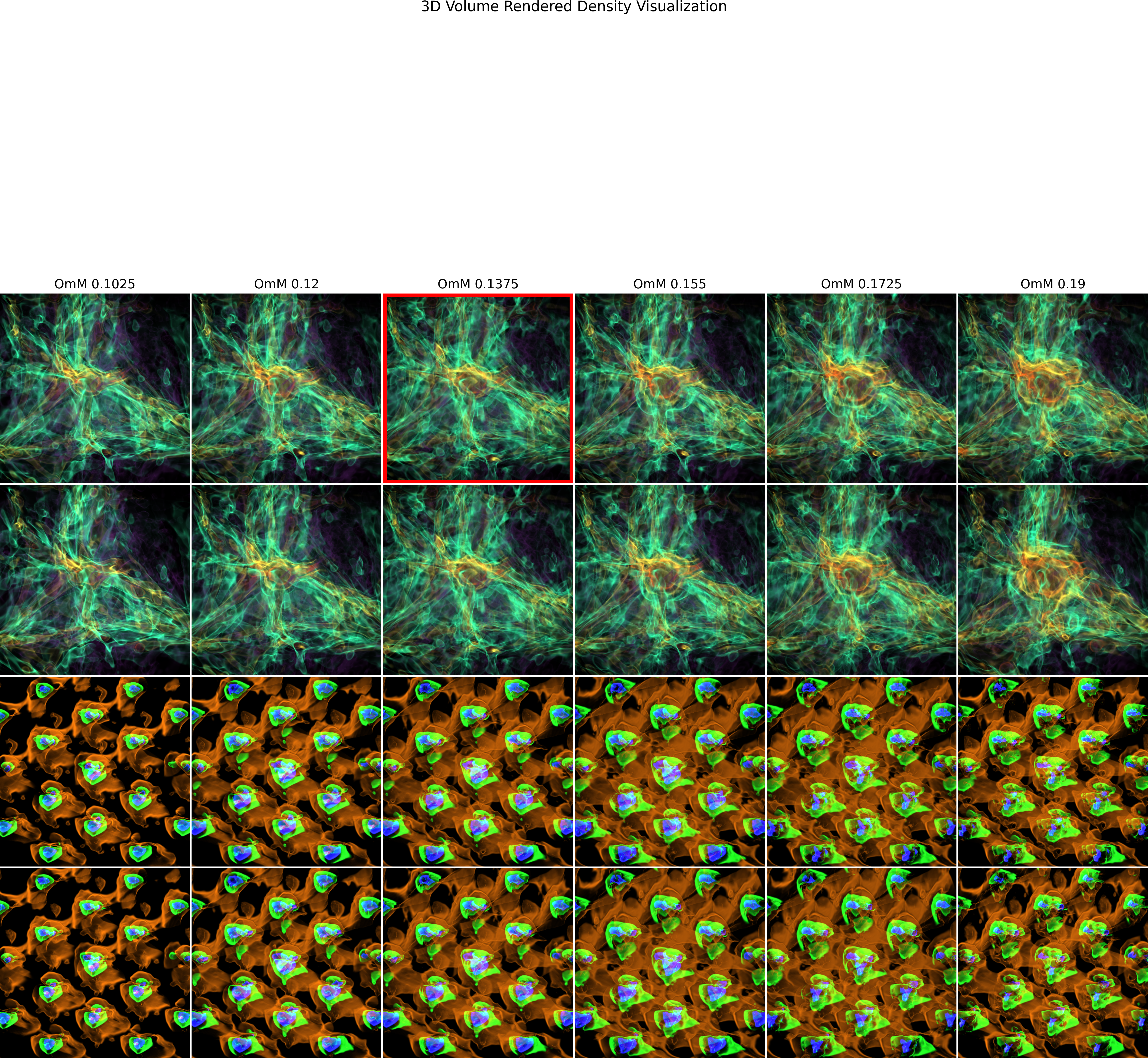}
            };
        \end{tikzpicture}
    \end{minipage}
    \hspace{5pt}
    \begin{minipage}[t]{0.15\linewidth}
    \vspace*{-275pt}
        \centering
        \includegraphics[trim=40 0 10 10, clip, width=\linewidth]{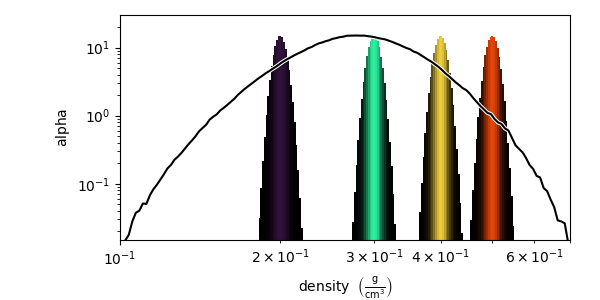}
        \vspace*{-20pt}
        \caption*{\begin{new}\footnotesize TF for density (black line---data distribution, Gaussian lobes---color and opacity.\end{new}}
        \vspace{10pt}
        \includegraphics[trim=40 0 10 10, clip, width=\linewidth]{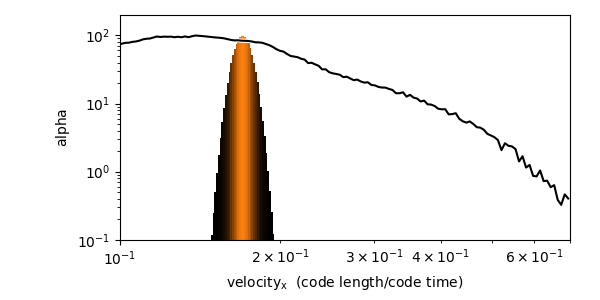}
        \vspace*{-20pt}
        \caption*{\begin{new}\footnotesize TF for flow $x$.\end{new}}
        \vspace{10pt}
        \includegraphics[trim=40 0 10 10, clip, width=\linewidth]{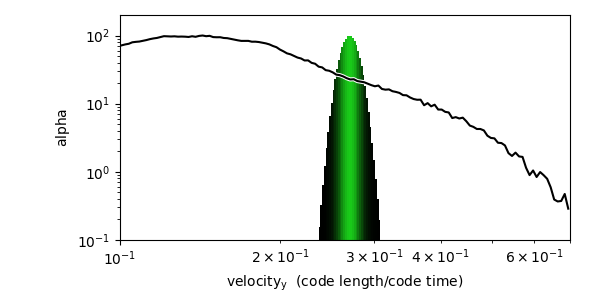}
        \vspace*{-20pt}
        \caption*{\begin{new}\footnotesize TF for flow $y$.\end{new}}
        \vspace{10pt}
        \includegraphics[trim=40 0 10 10, clip, width=\linewidth]{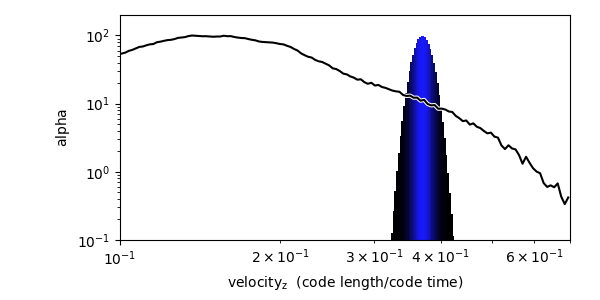}
        \vspace*{-20pt}
        \caption*{\begin{new}\footnotesize TF for flow $z$.\end{new}}
    \end{minipage}
    }
    \vspace{-12pt}
    \caption{Nyx simulation parameter space exploration \begin{new}and transfer functions for density and flow field components.\end{new}}
    \label{fig:nx_param_exp}
    \vspace{-18pt}
\end{figure*}
\end{old}

\begin{figure}[t!]
\scalebox{0.99}{
    \centering
    \includegraphics[trim=40 10 40 53, clip, width=\linewidth]{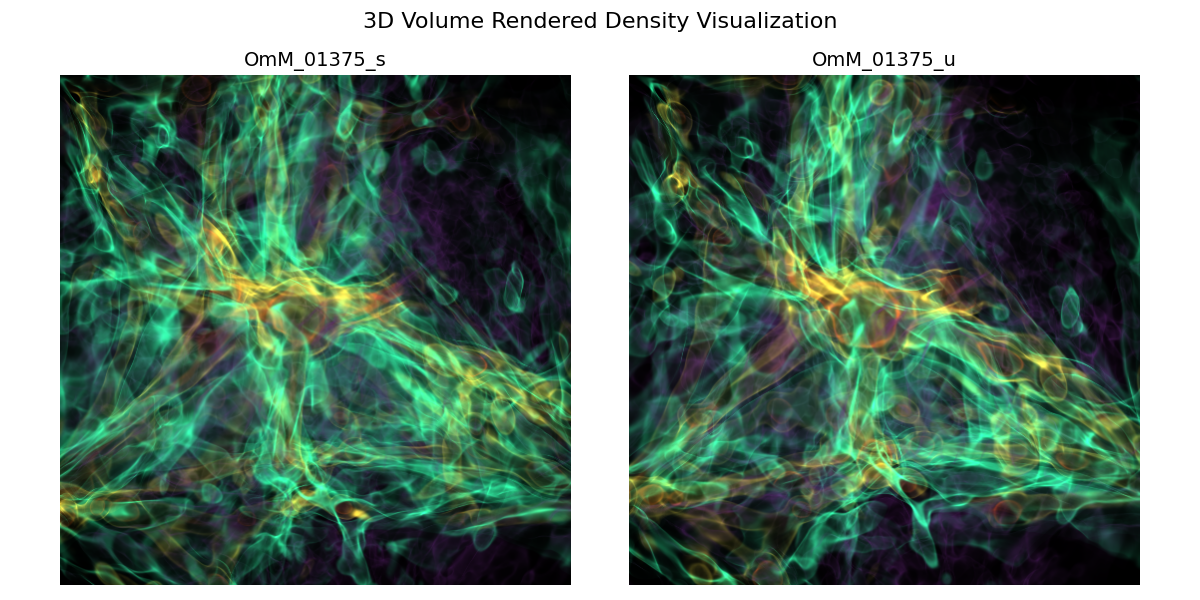}
    }
        \caption{\begin{new}The rendering of two volumes that were used for Nyx simulation parameter space exploration.\end{new}}
        \vspace{-12pt}
    \label{fig:ends_plot}
    \vspace{-12pt}
\end{figure}

\begin{figure*}[ht!]
    \centering
    \scalebox{0.95}{
    \begin{minipage}[t]{0.03\linewidth}
        \vspace*{-390pt}
        \text{\rotatebox{90}{\footnotesize Dens GT}} \\ [23pt]
        \text{\rotatebox{90}{\footnotesize Dens \acronym}} \\ [21pt]
        \text{\rotatebox{90}{\footnotesize \begin{new}Dens diff\end{new}}} \\ [32pt]
        \text{\rotatebox{90}{\footnotesize Flow GT}} \\ [28pt]
        \text{\rotatebox{90}{\footnotesize Flow \acronym}} \\ [23pt]
        \text{\rotatebox{90}{\footnotesize \begin{new}Flow diff\end{new}}} \\
    \end{minipage}
    \hspace*{3pt}
    \begin{minipage}[t]{0.82\linewidth}
        \begin{tikzpicture}
            \node (mainfig) at (0,0) {
                \includegraphics[trim=0 0 0 0, clip, width=0.98\linewidth]{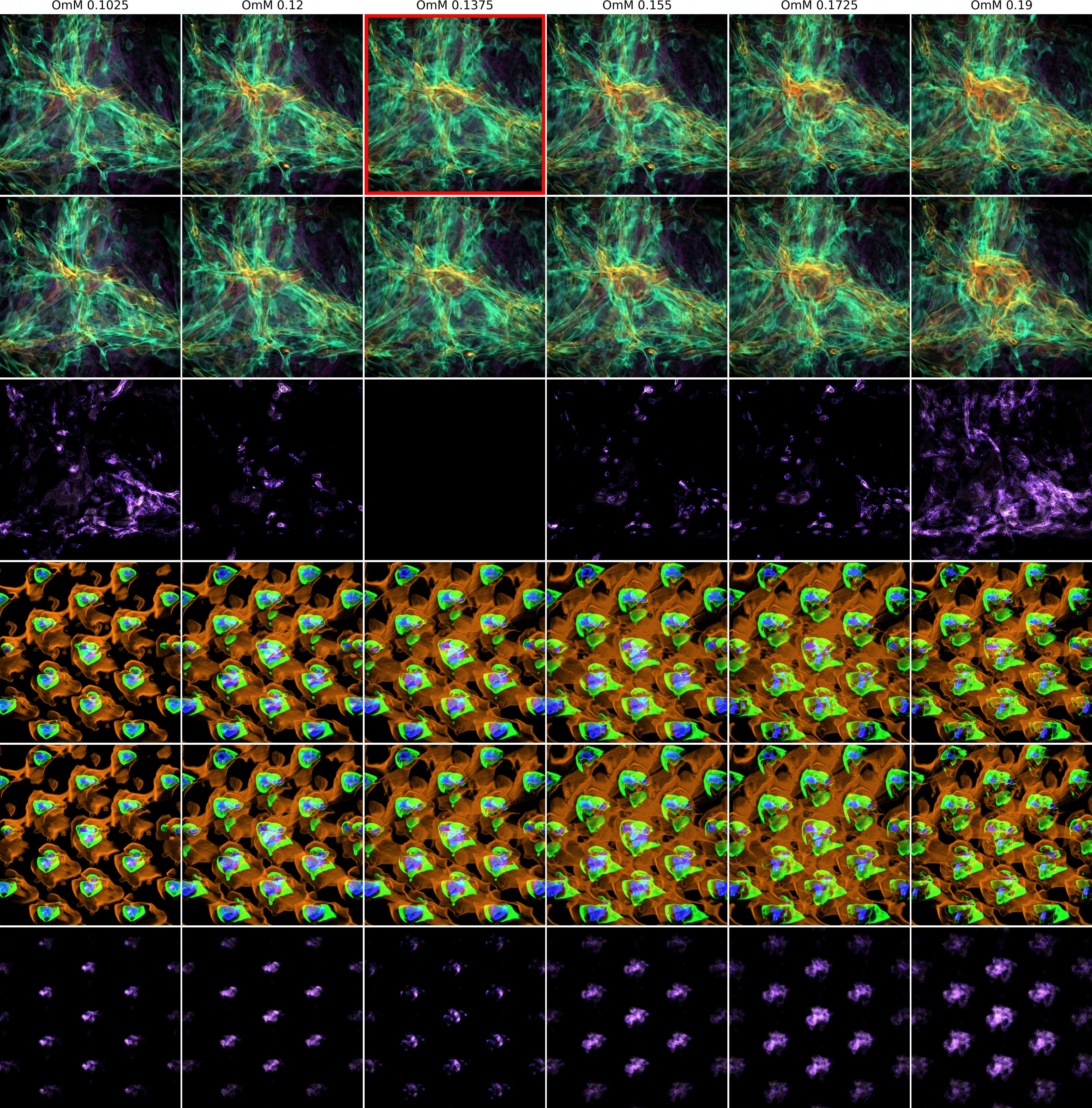}
            };
        \end{tikzpicture}
    \end{minipage}
    \hspace{5pt}
    \begin{minipage}[t]{0.15\linewidth}
    \vspace*{-385pt}
        \centering
        \includegraphics[trim=40 0 10 10, clip, width=\linewidth]{figures/Results/nyx3d/dens_gt_tf.png}
        \vspace*{-20pt}
        \caption*{\begin{new}\footnotesize TF for density (black line---data distribution, Gaussian lobes---color and opacity.\end{new}}
        \vspace{10pt}
        \includegraphics[trim=40 0 10 10, clip, width=\linewidth]{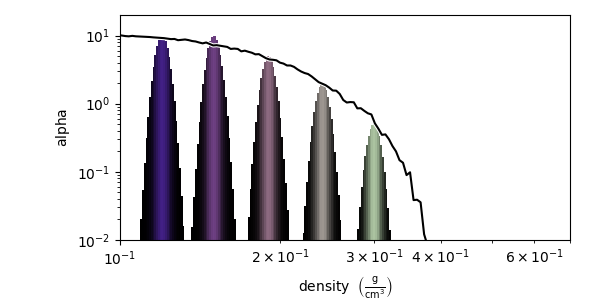}
        \vspace*{-20pt}
        \caption*{\begin{new}\footnotesize TF for density difference.\end{new}}
        \vspace{5pt}
        \includegraphics[trim=40 0 10 10, clip, width=\linewidth]{figures/Results/nyx3d/flow_x_gt_tf.png}
        \vspace*{-22pt}
        \caption*{\begin{new}\footnotesize TF for flow $x$.\end{new}}
        \vspace{2pt}
        \includegraphics[trim=40 0 10 10, clip, width=\linewidth]{figures/Results/nyx3d/flow_y_gt_tf.png}
        \vspace*{-22pt}
        \caption*{\begin{new}\footnotesize TF for flow $y$.\end{new}}
        \vspace{2pt}
        \includegraphics[trim=40 0 10 10, clip, width=\linewidth]{figures/Results/nyx3d/flow_z_gt_tf.png}
        \vspace*{-22pt}
        \caption*{\begin{new}\footnotesize TF for flow $z$.\end{new}}
        \vspace{2pt}
        \includegraphics[trim=40 0 10 10, clip, width=\linewidth]{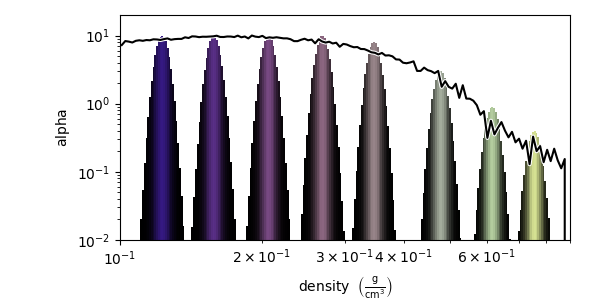}
        \vspace*{-22pt}
        \caption*{\begin{new}\footnotesize TF for flow difference.\end{new}}
    \end{minipage}
    }
    \vspace{-3pt}
    \caption{Nyx simulation parameter space exploration \begin{new}and transfer functions for density and flow field components. From top to bottom, the rows show GT density, \acronym interpolated density, difference between GT and \acronym density, GT flow, \acronym flow estimation, and difference between GT and \acronym flow.\end{new}}
    \label{fig:nx_param_exp}
    \vspace{-12pt}
\end{figure*}

\vspace{-10pt}
\subsection{Parameter-Driven Data Synthesis}
\vspace{-3pt}
\label{subsec:param_data}


Leveraging the HyperNet architecture, which conditions the FLINT* network on specific simulation parameters, \acronym can dynamically adapt its outputs based on changes in simulation settings.
\begin{old}
As illustrated in~\autoref{fig:nx_param_exp}, by providing two volumes at timesteps $s$ and $u$ (see \autoref{fig:ends_plot} of the supplementary material, showing renderings of these two volumes) around the target interpolation point (marked with a red outline) but varying only the simulation parameters through HyperNet, we can generate outputs that visually align with the ensemble volumes with those specific parameter values.
\end{old}
\begin{new}
As shown in~\autoref{fig:ends_plot}, which presents renderings of the input volumes at timesteps $s$ and $u$, these volumes serve as the foundation for interpolating the target timestep while varying only the simulation parameters through HyperNet. As illustrated in~\autoref{fig:nx_param_exp}, this approach effectively generates outputs that closely match the ensemble volumes associated with the specified parameter values.
\end{new}
Even when starting with identical initial data, \acronym effectively adapts its predictions solely based on the altered parameter inputs. 
For instance, the density reconstruction and flow estimation are particularly accurate for the parameter setting $\Omega_m = 0.1375$ (third column), corresponding to the data from the ensemble member used as inputs.
Notably, \acronym maintains the structural integrity and fine-grained details of the density field (second row) while accurately capturing the dynamic flow patterns (\begin{new}fifth\end{new} row), demonstrating its capability to preserve both shape and texture.
By using the same input volume and altering only the simulation parameters, \acronym generates new data outputs effectively for settings like $\Omega_m = 0.12$, $0.155$, and $0.1725$, achieving visually close reconstructions, \begin{new}as the difference plots confirm\end{new}, showcasing its generalization capabilities. 
\begin{new}
While novel (unseen) simulation features cannot be generated, these results suggest that the model is capable of capturing parameter-dependent relationships, as reflected in the smooth density transitions and flow field predictions that generally follow GT dynamic trends.
\end{new}
The reconstruction remains robust despite these novel parameters.
As expected, we note a slight increase in error for more distant parameter values such as $\Omega_m = 0.1025$ and $0.19$ (first and last columns, respectively). 
This is likely due to these configurations being further away from the range of parameters the model received as input during inference. 
Nonetheless, the results remain decent, with distinct features preserved, such as central galaxy density formation and distinct flow patterns, showcasing \acronym's parameter-dependent 
generalization capabilities.

\vspace{-10pt}
\section{Conclusion and Future Work}
\vspace{-2pt}

In this work, we have proposed \acronym, a hypernetwork-based method for flow estimation and scalar field interpolation in spatio-temporal scientific ensembles. By leveraging hypernetworks, \acronym adapts to varying simulation parameters, enabling accurate flow estimation and high-quality temporal interpolants, even with limited data. It outperforms recent state-of-the-art methods while achieving fast inference without requiring extensive pre-training or fine-tuning on simplified datasets. Validated across diverse simulation ensembles, \acronym demonstrates robust and versatile performance, making it a valuable tool for scientific visualization.


\vspace{-3pt}
The integration of hypernetworks in \acronym establishes a robust framework for analyzing how simulation parameters influence flow and density dynamics, enabling efficient parameter space exploration and deeper insights into complex systems. 
\begin{new}Integrating skill score metrics that account for spatial and temporal biases, along with adaptivity based on flow characteristics and simulation constraints, could enhance robustness and applicability in future work.\end{new}
Incorporating stable diffusion techniques~\cite{rombach2022high} could further enhance \acronym's ability to generate high-quality vector and scalar fields by leveraging simulation parameters. Known for producing smooth outputs, stable diffusion methods could complement hypernetworks, improving generalization in intricate simulations.

\begin{new}
\vspace{-10pt}
\section*{Acknowledgments}
\vspace{-2pt}
We thank the Center for Information Technology of the University of Groningen for their support and for providing access to the Hábrók high performance computing cluster. 
We also thank the Deutsche Forschungsgemeinschaft (DFG, German Research Foundation) for supporting this work by funding SFB 1313 (Project Number 327154368).
We thank dr.\ Maxime Trebitsch (Institut d’Astrophysique de Paris) for his help in interpreting the flow visualizations of the Nyx cosmological simulation and assessing their usefulness for data analysis.
\end{new}

\bibliographystyle{eg-alpha-doi} 
\bibliography{EGauthorGuidelines-cgf-fin}

\onecolumn
\newpage
\twocolumn

\appendix
\begin{new}
\section{\textbf{Supplementary Material}}
\end{new}

\vspace{6pt}
\subsection{Nyx and Castro Transfer Functions}
\label{subsec:tf}

\begin{figure}[h]
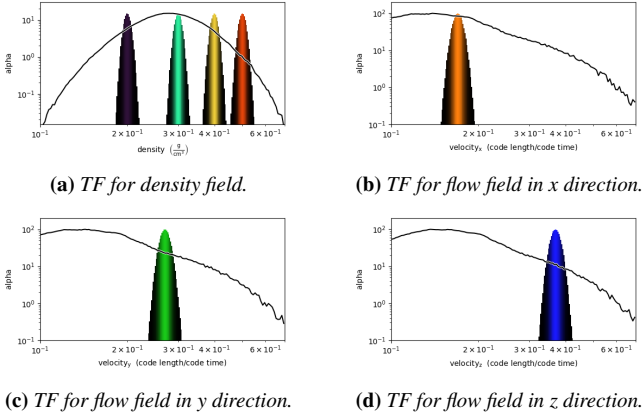

    \centering
    \begin{subfigure}{0.45\linewidth}
        \centering
        \includegraphics[trim=40 0 10 10, clip, width=\linewidth]{figures/Results/nyx3d/dens_gt_tf.png}
        \caption{TF for density field.}
        \label{fig:density_tf_nyx}
        \vspace{6pt}
    \end{subfigure}
    \hfill
    \begin{subfigure}{0.45\linewidth}
        \centering
        \includegraphics[trim=40 0 10 10, clip, width=\linewidth]{figures/Results/nyx3d/flow_x_gt_tf.png}
        \caption{TF for flow field in $x$ direction.}
        \label{fig:flow_x_tf_nyx}
        \vspace{6pt}
    \end{subfigure}
    \begin{subfigure}{0.45\linewidth}
        \centering
        \includegraphics[trim=40 0 10 10, clip, width=\linewidth]{figures/Results/nyx3d/flow_y_gt_tf.png}
        \caption{TF for flow field in $y$ direction.}
        \label{fig:flow_y_tf_nyx}
    \end{subfigure}
    \hfill
    \begin{subfigure}{0.45\linewidth}
        \centering
        \includegraphics[trim=40 0 10 10, clip, width=\linewidth]{figures/Results/nyx3d/flow_z_gt_tf.png}
        \caption{TF for flow field in $z$ direction.}
        \label{fig:flow_z_tf_nyx}
    \end{subfigure}
    \caption{Nyx ensemble: transfer function for density and $x$, $y$, $z$ components of the flow field.}
    \label{fig:combined_tf_nyx}
\end{figure}

\begin{figure}[H]
    \centering
    \begin{subfigure}{0.45\linewidth}
        \centering
        \includegraphics[trim=40 0 10 10, clip, width=\linewidth]{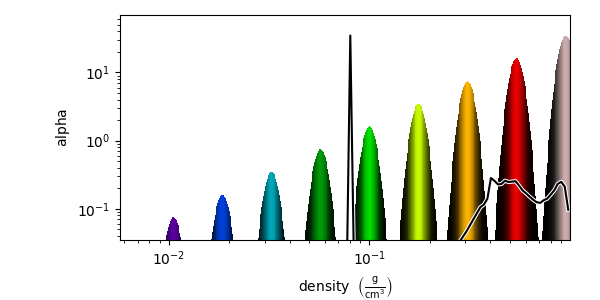}
        \caption{TF for density field.}
        \label{fig:density_tf_castro}
        \vspace{6pt}
    \end{subfigure}
    \hfill
    \begin{subfigure}{0.45\linewidth}
        \centering
        \includegraphics[trim=40 0 10 10, clip, width=\linewidth]{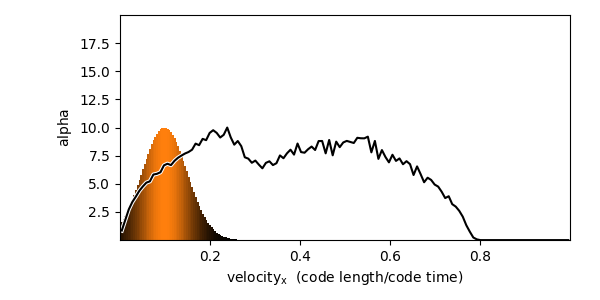}
        \caption{TF for flow field in $x$ direction.}
        \label{fig:flow_x_tf_castro}
        \vspace{6pt}
    \end{subfigure}
    
    \begin{subfigure}{0.45\linewidth}
        \centering
        \includegraphics[trim=40 0 10 10, clip, width=\linewidth]{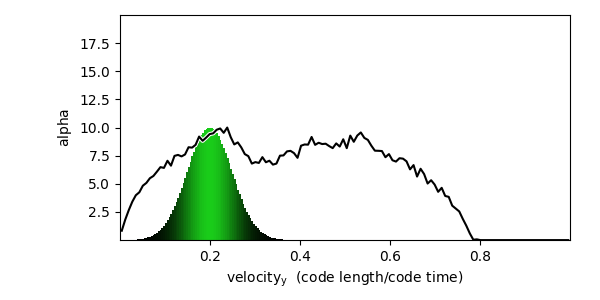}
        \caption{TF for flow field in $y$ direction.}
        \label{fig:flow_y_tf_castro}
    \end{subfigure}
    \hfill
    \begin{subfigure}{0.45\linewidth}
        \centering
        \includegraphics[trim=40 0 10 10, clip, width=\linewidth]{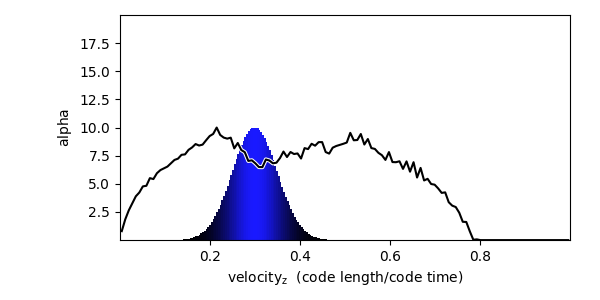}
        \caption{TF for flow field in $z$ direction.}
        \label{fig:flow_z_tf_castro}
    \end{subfigure}
    \caption{Castro ensemble: transfer function for density and $x$, $y$, $z$ components of the flow field.}
    \label{fig:combined_tf_castro}
\end{figure}

In \autoref{fig:combined_tf_nyx} and \autoref{fig:combined_tf_castro} we present transfer functions for visualizing the density and velocity components of the Nyx and Castro ensembles, respectively. 
These are applied to the density field and the $x$, $y$, and $z$ components of the flow field, offering an intuitive means to explore and analyze the spatial and dynamic characteristics of the simulation data. 
The utilized transfer functions help to better understand the structural patterns and flow dynamics inherent to each ensemble.

\begin{old}
\acronym takes as input the simulation parameters (for HypetNet), two scalar fields $D_s$ and $D_{u}$ of the same ensemble member at timesteps $s<u$, and an intermediate timestep $t$, where $s < t < u$ (for FLINT*).  
The goal is to predict the corresponding flow field $\hat{F}_{t}$ and generate interpolated scalar fields $\hat{D}_{t}$ for any intermediate time $t\in [s,u]$.
To accomplish this, FLINT* first computes intermediate flow fields, $\Fts{}$ and $\Ftu{}$. The \emph{time-backward} flow field, $\Fts{}$, represents the flow vectors from the frame at time $t$ to an earlier frame at $s$. Conversely, the \emph{time-forward} flow, $\Ftu{}$, represents flow vectors from the frame at $t$  to a later frame at $u$. These intermediate flow fields are then used to warp scalar fields towards the target time $t$, generating estimates of the scalar field for that time step.
In the final step, \acronym combines the intermediate warped scalar fields using a fusion mask $M$ learned by FLINT*, where $M(i, j) \in [0, 1],\forall i, j$, which ensures smooth blending and high-quality interpolation. 
\end{old}

\subsection{3D Flow Estimation and Density Interpolation Results}
\label{subsec:results_figs}

\begin{figure*}[!t]
    \centering
    \begin{minipage}[t]{0.03\linewidth}
        \vspace*{-560pt}
        \text{\rotatebox{90}{\footnotesize Dens GT}} \\ [35pt]
        \text{\rotatebox{90}{\footnotesize Dens \acronym}} \\ [33pt]
        \text{\rotatebox{90}{\footnotesize Dens FLINT}} \\ [33pt]
        \text{\rotatebox{90}{\footnotesize Dens STSR-INR}} \\ [42pt]
        \text{\rotatebox{90}{\footnotesize Flow GT}} \\ [38pt]
        \text{\rotatebox{90}{\footnotesize Flow \acronym}} \\ [33pt]
        \text{\rotatebox{90}{\footnotesize Flow FLINT}} \\
    \end{minipage}
    \hspace*{3pt}
    \begin{minipage}[t]{0.95\linewidth}
        \begin{tikzpicture}
            \node (mainfig) at (0,0) {
                \includegraphics[trim=567 62 536 62, clip, width=0.93\linewidth]{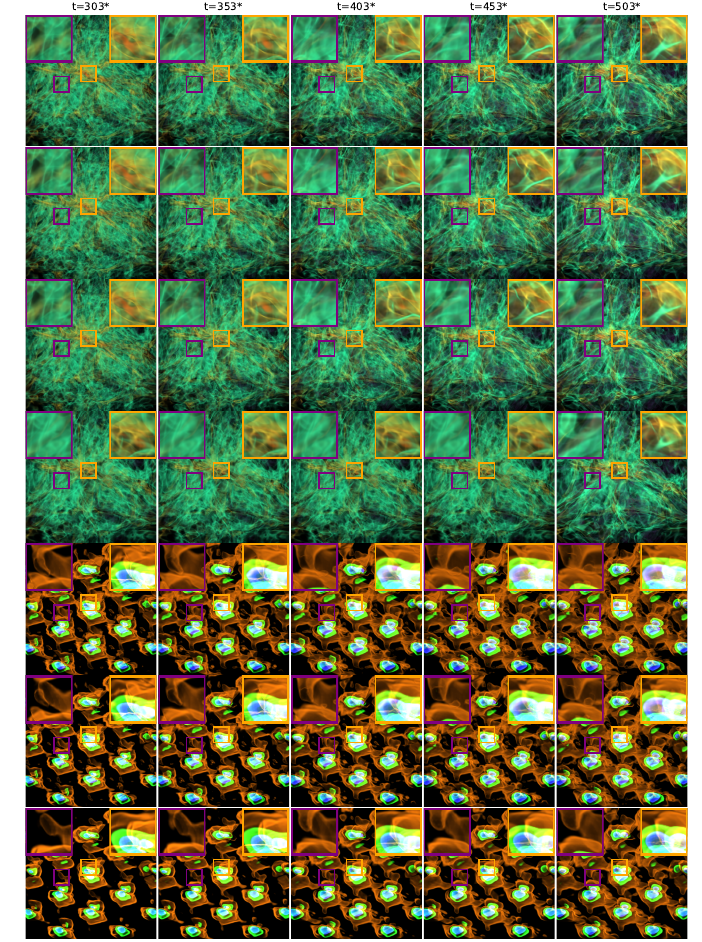}
            };
        \end{tikzpicture}
    \end{minipage}
    \caption{Nyx: \acronym flow field estimation and temporal density interpolation, 5$\times$.
        From top to bottom, the rows show GT density, \acronym interpolated density, FLINT interpolation, STSR-INR interpolation, GT flow, \acronym flow estimation, and FLINT flow estimation. 
        3D rendering was used for the density and flow visualization (\protect\orangecircle{} \protect\greencircle{} \protect\bluecircle{} colors representing $x$, $y$, and $z$ flow directions respectively).
        \vspace{-12pt}
        }
    \label{fig:nyx}
\end{figure*}

\begin{figure*}[!t]
    \centering
    \begin{minipage}[t]{0.03\linewidth}
        \vspace*{-560pt}
        \text{\rotatebox{90}{\footnotesize Dens GT}} \\ [35pt]
        \text{\rotatebox{90}{\footnotesize Dens \acronym}} \\ [33pt]
        \text{\rotatebox{90}{\footnotesize Dens FLINT}} \\ [33pt]
        \text{\rotatebox{90}{\footnotesize Dens STSR-INR}} \\ [42pt]
        \text{\rotatebox{90}{\footnotesize Flow GT}} \\ [38pt]
        \text{\rotatebox{90}{\footnotesize Flow \acronym}} \\ [33pt]
        \text{\rotatebox{90}{\footnotesize Flow FLINT}} \\
    \end{minipage}
    \hspace*{3pt}
    \begin{minipage}[t]{0.95\linewidth}
        \begin{tikzpicture}
            \node (mainfig) at (0,0) {
                \includegraphics[trim=567 62 536 62, clip, width=0.93\linewidth]{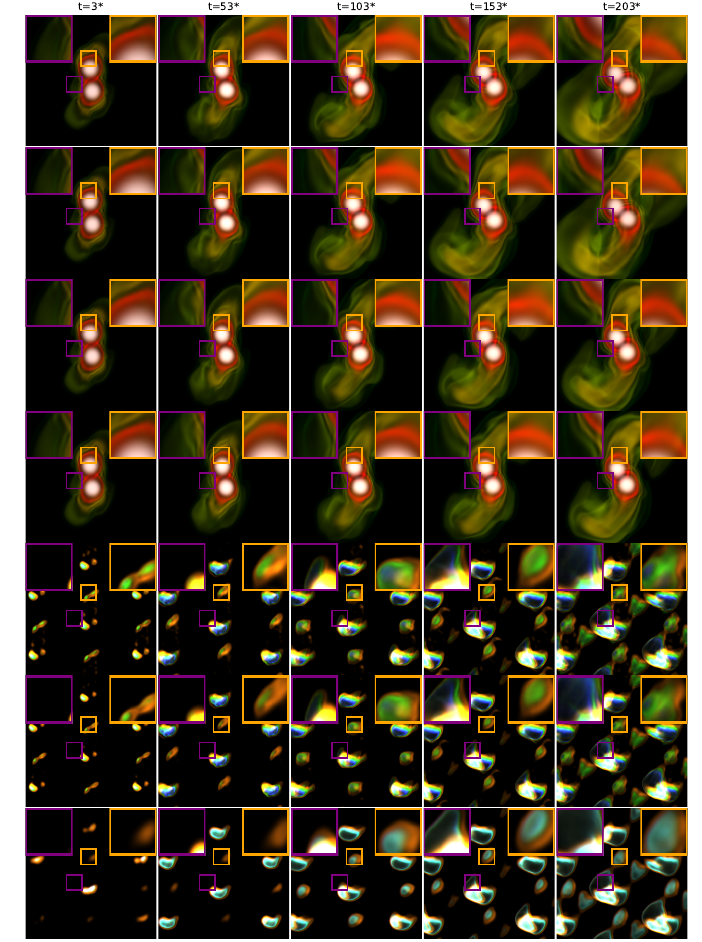}
            };
        \end{tikzpicture}
    \end{minipage}
    \caption{Castro: \acronym flow field estimation and temporal density interpolation, 5$\times$.
        From top to bottom, the rows show GT density, \acronym interpolated density, FLINT interpolation, STSR-INR interpolation, GT flow, \acronym flow estimation, and FLINT flow estimation. 
        3D rendering was used for the density and flow visualization (\protect\orangecircle{} \protect\greencircle{} \protect\bluecircle{} colors representing $x$, $y$, and $z$ flow directions respectively).
        \vspace{-12pt}
        }
    \label{fig:castro}
\end{figure*}

This subsection presents the complete results for 3D flow estimation and temporal interpolation tasks on the Nyx (\autoref{fig:nyx}) and Castro (\autoref{fig:castro}) datasets. The figures include visualizations of the estimated flow fields and interpolated density fields.
Additionally, the results highlight comparisons with baseline methods, including FLINT and STSR-INR.

\begin{new}
The 3D PSNR metric used in our evaluation is computed after normalizing the scalar data to [0,1], using the following formulation:
\begin{equation}
    \text{MSE} = \frac{1}{m n o} \sum_{i=0}^{m-1} \sum_{j=0}^{n-1} \sum_{k=0}^{o-1} (D^{GT}_{t,i,j,k} - \hat{D}_{t,i,j,k})^2,
\end{equation}
\vspace{-6pt}
\begin{equation}
    \text{PSNR} = 20 \log_{10} (1.0) - 10 \log_{10} (\text{MSE}),
\end{equation}
where the mean squared error (MSE) is calculated over the entire 3D volume, with \( m, n, o \) representing the spatial dimensions of the volume along the \( x \)-, \( y \)-, and \( z \)-axes, respectively.

For flow estimation, we compute the 3D endpoint error (EPE) as:
\begin{equation}
    \text{EPE} = \frac{1}{m n o} \sum_{i=0}^{m-1} \sum_{j=0}^{n-1} \sum_{k=0}^{o-1} \left\lVert F^{GT}_{t,i,j,k} - \hat{F}_{t,i,j,k} \right\rVert_2.
\end{equation}
Here, \( F^{GT} \) and \( \hat{F} \) represent the ground-truth and predicted flow fields, respectively, with components along the \( x \)-, \( y \)-, and \( z \)-axes. The EPE measures the Euclidean distance between the predicted and true flow vectors across the entire 3D volume.

\end{new}

\end{document}